%% file: neurips_2025.tex
\documentclass{article}

\usepackage[preprint]{neurips_2025}
\usepackage[pagebackref,breaklinks,colorlinks,citecolor=cyan,linkcolor=cyan]{hyperref}

\usepackage[utf8]{inputenc} 
\usepackage[T1]{fontenc}    
\usepackage{hyperref}       
\usepackage{url}            
\usepackage{booktabs}       
\usepackage{amsfonts}       
\usepackage{nicefrac}       
\usepackage{microtype}      
\usepackage{xcolor}         

\usepackage{color, colortbl}  
\usepackage{graphicx}
\usepackage{enumitem}  

\usepackage{multirow}
\usepackage{makecell}
\usepackage{wrapfig}
\usepackage{tikz}

\usepackage{amsmath}
\usepackage{amssymb}
\usepackage{algorithm}
\usepackage{algorithmic}
\usepackage{bm}
\usepackage{bbold}

\usepackage{pifont} 
\newcommand{\cmark}{\ding{51}}
\newcommand{\xmark}{\ding{55}}

\definecolor{LightCyan}{rgb}{0.88,1,1}

\newcommand{\best}[1]{{\textbf{#1}}}
\newcommand{\second}[1]{{\underline{#1}}}
\newcommand{\up}[1]{\textcolor{red}{\scriptsize (+#1)}}
\newcommand{\down}[1]{\textcolor[RGB]{46,139,87}{\scriptsize (-#1)}}
\newcommand{\tie}[1]{\textcolor{gray}{\scriptsize (-)}}
\newcommand{\relativeimprove}[1]{\textit{+#1\%}}
\newcommand{\gray}[1]{\textcolor{gray}{#1}}

\usepackage{gradient-text}  
\usepackage{xspace}
\newcommand\modelrgb{\textsc{\gradientRGB{Unite}{72, 203, 194}{72, 116, 203}}\xspace}
\newcommand\model{\textsc{Unite}\xspace}
\newcommand\modelbase{\textsc{Unite}$_\text{base}$\xspace}
\newcommand\modelinstruct{\textsc{Unite}$_\text{instruct}$\xspace}
\newcommand\modalmask{MAMCL\xspace}

\definecolor{Magenta}{rgb}{0.8, 0.1, 0.6}

\title{Modality Curation: Building Universal Embeddings for Advanced Multimodal Information Retrieval}

\author{
Fanheng Kong\textsuperscript{1,2*$\ddagger$}, Jingyuan Zhang\textsuperscript{2*},
Yahui Liu\textsuperscript{2}, Hongzhi Zhang\textsuperscript{2}, Shi Feng\textsuperscript{1$\dagger$}, \\
\textbf{Xiaocui Yang}\textsuperscript{1}, \textbf{Daling Wang}\textsuperscript{1},
\textbf{Yu Tian}\textsuperscript{2}, \textbf{Victoria W.}, \textbf{Fuzheng Zhang}\textsuperscript{2}, \textbf{Guorui Zhou}\textsuperscript{2} \\
\textsuperscript{1}Northeastern University \quad
\textsuperscript{2}Kuaishou Technology \\
\tt\small kongfanheng426@gmail.com, \tt\small fengshi@cse.neu.edu.cn
}

\begin{document}

\maketitle 

\input{sections/0-abstract}

{
\renewcommand{\thefootnote}%
{\fnsymbol{footnote}}
\footnotetext[0]{* Equal contributions. $\dagger$ Corresponding author. $\ddagger$ Work done during an internship at Kuaishou Technology.} 
}


\vspace{-1em}
\begin{figure}[h]
    \centering
    \includegraphics[width=0.98\linewidth]{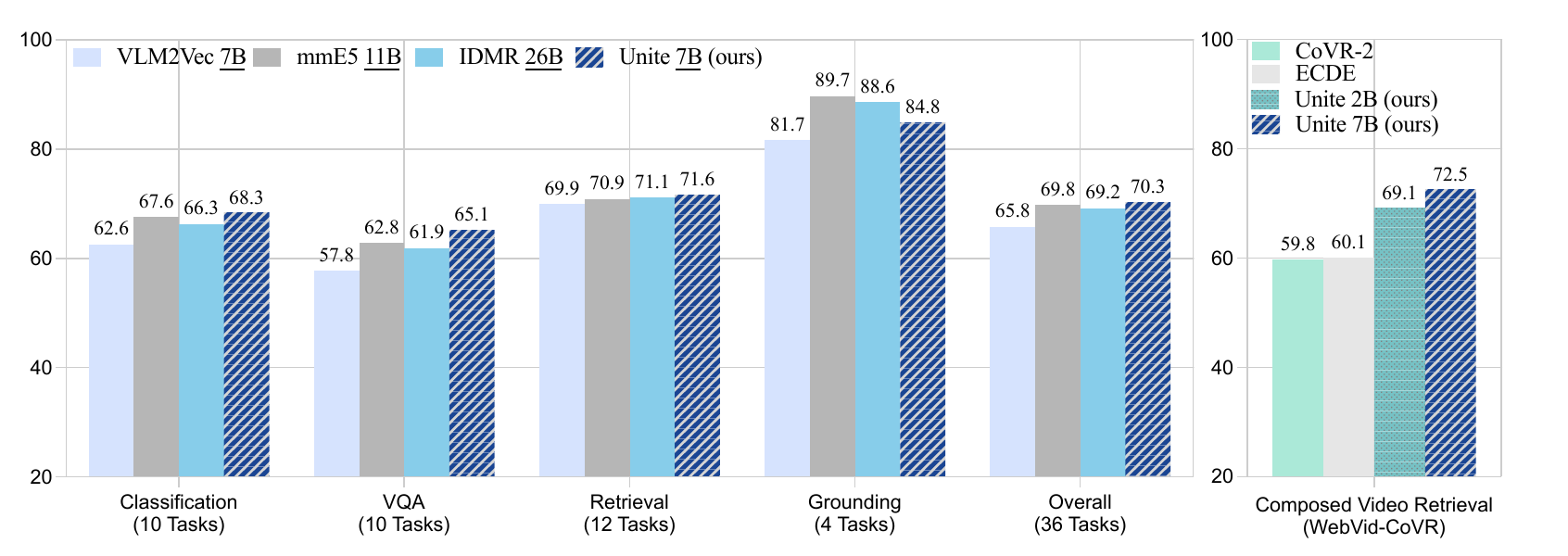}
    \caption{Performance comparison on instruction-based retrieval benchmarks (left: MMEB~\cite{jiang2024vlm2vec} and right: WebVid-CoVR~\cite{ventura2024covr}). Our \modelrgb achieves leading performance on various tasks, even surpassing models with larger parameter scales.}
    \label{fig:bar_instruct}
    \vspace{-1em}
\end{figure}

\input{sections/1-introduction}

\input{sections/2-related-work}
\input{sections/3-method}
\input{sections/4-experiments}

\input{sections/5-conclusion}

\clearpage
\bibliographystyle{plain}
\bibliography{cite}

\clearpage
\input{sections/appendix}

\end{document}

%% file: sections/0-abstract.tex
\begin{abstract}

Multimodal information retrieval (MIR) faces inherent challenges due to the heterogeneity of data sources and the complexity of cross-modal alignment. While previous studies have identified modal gaps in feature spaces, a systematic approach to address these challenges remains unexplored.
In this work, we introduce \modelrgb, a universal framework that tackles these challenges through two critical yet underexplored aspects: \textit{data curation} and \textit{modality-aware training configurations}. Our work provides the first comprehensive analysis of how modality-specific data properties influence downstream task performance across diverse scenarios. Moreover, we propose Modal-Aware Masked Contrastive Learning (MAMCL) to mitigate the competitive relationships among the instances of different modalities.
Our framework achieves state-of-the-art results on multiple multimodal retrieval benchmarks, outperforming existing methods by notable margins.
Through extensive experiments, we demonstrate that strategic modality curation and tailored training protocols are pivotal for robust cross-modal representation learning.
This work not only advances MIR performance but also provides a foundational blueprint for future research in multimodal systems.
Our project is available at \url{https://friedrichor.github.io/projects/UNITE}.

\end{abstract}

%% file: sections/1-introduction.tex
\section{Introduction}
\label{sec:introduction}

Multimodal Information Retrieval (MIR) is a critical research topic~\cite{radford2021clip,jia2021align}, aiming to satisfy users' information requirements for diverse media, such as text, images, and videos. 
As multimedia applications continue to progress and develop, a series of more complex and demanding tasks come to the fore, which are collectively referred to as fused-modal retrieval such as the retrieval of composite images or videos~\cite{liu2021cirr,zhang2024magiclens,ventura2024covr}. 
These tasks require highly sophisticated approaches for handling interleaved multimodal queries and candidates, highlighting the necessity for the development of a one-piece framework for unified multimodal representations. 

Recently, large multimodal models (LMMs) have shown powerful capabilities on various vision-language tasks, such as visual question answering (VQA)~\cite{antol2015vqa, kafle2017visual, wang2017fvqa, marino2019okvqa} and multimodal fact-checking~\cite{yao2023end, akhtar2023multimodal}. 
In MIR field, several methods~\cite{jiang2024vlm2vec,liu2024lamra,gu2025unime} have explored adapting large language models (LLMs) for retrieval tasks via contrastive learning, aiming to produce unified embeddings. 
For example, E5-V~\cite{jiang2024e5v} finetunes LLaVA-NeXT~\cite{liu2024llavanext} with text-only NLI~\citep{gao2021simcse} data, demonstrating the portability of LMMs for multimodal retrieval. 
GME~\citep{zhang2024gme} achieves leading performance in various image-text retrieval by finetuning Qwen2-VL~\cite{wang2024qwen2vl} on diverse image-text datasets.
InternVideo2~\cite{wang2024internvideo2} stands out prominently in text-video retrieval, due to its training process that involves on hundred millions of video-text pairs. 
Although achieving notable success in specific domains, these models are hindered by their limited modalities, which inherently restrict their ability to fully capitalize on the potential of LMMs for generating unified multimodal embeddings. 

Despite ongoing researches~\cite{mckinzie2024mm1, zhang2024mm1.5, zohar2024apollo} exploring training strategies for LMMs in MIR--including model architectures, training methodologies, and dataset considerations--critical questions remain unresolved. 
Specifically, the optimal data composition and proportions, as well as the nuanced impact of different modal data configurations across various retrieval tasks, have yet to be comprehensively understood.
In our empirical investigations, we find that inappropriate combinations of multimodal data or data training sequences can easily disrupt the harmonious integration of diverse data modalities,  causing the model to misinterpret the relationships between different types of information.

In this paper, through a meticulous analysis of how various data compositions impact retrieval results, we make efforts to achieve a balance among the three modalities of text, image, and video. In particular, we find that introducing a small number of fine-grained video-text pairs during the retrieval adaptation stage can significantly enhance the fine-grained retrieval performance of LMMs. Meanwhile, existing works~\cite{jiang2024e5v,zhang2024gme} have proven that there is a substantial distribution gap in the data of various modalities within the feature space. Simply mixing data from different modalities for contrastive learning will affect the quality of representation learning (\textit{i.e.}, introducing noise). Therefore, we propose \textit{Modal-Aware Masked Contrastive Learning} (MAMCL) to balance the competitive relationships among the instances of various modalities.

Finally, we develop a \textbf{UNI}versal mul\textbf{T}imodal \textbf{E}mbedder, named \modelrgb, that effectively handles text, images, videos, and their combinations. During the training, we employ an evolving training strategy to gradually unlock the retrieval capability of LMMs. After two training stages (\textit{i.e.}, retrieval adaptation and instruction tuning), we validate our approach through comprehensive evaluation on 40+ diverse retrieval tasks, spanning coarse-grained, fine-grained, and instruction-based retrieval across text, images, and video. Experimental results show that \modelrgb achieves state-of-the-art performance in various tasks, and outperforms existing specialized domain-specific models in numerous scenarios. 

In summary, our contributions are as follows:
\begin{itemize}[leftmargin=*, nolistsep, noitemsep]
    \item We unveil the appropriate method for curating modality data during the process of learning unified multimodal embeddings, so as to balance the gap between the feature spaces of different modalities. 
    \item We propose MAMCL to mitigate the competitive relationships among the instances of various modalities. Specially, the MAMCL strategy can serve as a general method and be applied to any extended modal scenarios.
    \item To the best of our knowledge, \modelrgb is the first model capable of enabling text, image, video and fused-modal to concurrently focus on fine-grained and instruction-based retrieval tasks. Notably, \modelrgb achieves state-of-the-art performance in 40+ different tasks, surpassing specialized domain-specific models in numerous scenarios, as shown in Figure~\ref{fig:bar_instruct}.
\end{itemize}

%% file: sections/2-related-work.tex
\section{Related Work}
\label{sec:related_work}

\textbf{Large Multimodal Models.} LMMs have mushroomed, showcasing impressive capabilities in multimodal information understanding~\cite{li2023blip2,zhang2025capybaravl}.
Pioneering efforts such as LLaVA~\cite{liu2024llava, liu2024llava1.5}, MiniGPT-4~\cite{zhu2023minigpt4}, InternVL~\cite{chen2024internvl1.5,zhu2025internvl3}, and Qwen-VL~\cite{wang2024qwen2vl,bai2025qwen2.5vl}, bridge visual encoders and LLM with lightweight intermediate architectures. 
Recent advances~\cite{li2024llavaov,wang2025internvideo2.5,zohar2024apollo,xu2024pllava,maaz2024videogpt+, lin2023videollava} have extended these techniques from static images to sequential videos, demonstrating promising results in video understanding through frame-based processing. 
These developments have catalyzed the widespread adoption of LMMs across various applications~\cite{liu2024llavaplus,pan2023kosmosg,zhang2024stickerconv,kong2025tuna}.

\textbf{Cross-Modal Retrieval.} Traditional multimodal retrieval primarily focuses on cross-modal scenarios, particularly text-image retrieval~\cite{wang2016comprehensive,wang2025cross} and text-video retrieval~\cite{zhu2023deep,jiang2022hcmi}. 
Foundational methods, such as CLIP~\cite{radford2021clip}, Align~\cite{jia2021align}, BLIP~\cite{li2022blip} and CoCa~\cite{yu2022coca}, separately encode text and images through dual-encoder structures, and learn multimodal representations by contrastive learning on large-scale image-text pairs. 
ImageBind~\cite{girdhar2023imagebind}, OmniBind~\cite{wang2024omnibind} and VAST~\cite{chen2023vast} expand this approach to accommodate more modalities with similar architectures. 

\textbf{Instruction-based Retrieval.} Recently, the community has witnessed growing demand for multimodal information retrieval in complex scenarios, including composed image retrieval~\cite{liu2021cirr,saito2023pic2word,zhang2024magiclens}, composed video retrieval~\cite{ventura2024covr,hummel2024egocvr}, multimodal document retrieval~\cite{mathew2021docvqa,ma2024dse,dong2025mmdocir}, multimodal knowledge retrieval~\cite{luo2023end,long2024generative}, and multimodal retrieval-augmented generation~\cite{chen2022murag,zhao2023retrieving,caffagni2024wikillava,luo2024videorag}. 
While CLIP-based models struggle with complex multimodal queries and candidates, LMMs naturally excel at processing fused-modal inputs within a universal framework. 
Recent methods~\cite{zhang2024gme, liu2024lamra, lin2024mmembed, jiang2024vlm2vec, lan2025llave, liu2025idmr, lan2025llave} contrastively train LMMs (\textit{e.g.}, Qwen2-VL~\cite{wang2024qwen2vl} and LLaVA-NeXT~\cite{liu2024llavanext}) to uniformly embed images and text, boosting complex fused-modal retrieval. 
However, these models still face limitations in video-related tasks compared to video-specialized methods, highlighting the need for a universal multimodal embedder that excels across text, image, video, and their combinations.

%% file: sections/3-method.tex
\section{\model}
\label{sec:model}

\subsection{Task Formulation}
\label{subsec:task_formulation}

Unified multimodal retrieval addresses queries and candidates across text, image, and their combinations~\cite{wei2024uniir,zhang2024gme}. 
We extend prior formulations to include video, enabling more comprehensive multimodal embedding alignment. 
We define a query $\mathbf{q} \in \mathcal{Q} \subset \mathbb{R}^d$ representing an embedding that is sampled from a specific modality.
Specifically, $\mathbf{q}_t$, $\mathbf{q}_i$ and $\mathbf{q}_v$ denote the text, image and video embeddings, respectively.  
In practice, the query can be any single modality or a combination of various modalities (\textit{e.g.}, ($\mathbf{q}_t$, $\mathbf{q}_i$), ($\mathbf{q}_v$, $\mathbf{q}_t$)).
Similarly, we define a retrieval candidate $\mathbf{c} \in \mathcal{Q} \subset \mathbb{R}^{d}$ presenting the representation embeddings of a specific modality.
Thus, $\mathbf{c}_t$, $\mathbf{c}_i$ and $\mathbf{c}_v$ denote the text, image and video embeddings, respectively. 
The main purpose of MIR is to maximize the correlation between the related query $\mathbf{q}$ and candidate $\mathbf{c}$ pairs in the repsentation space $\mathcal{Q}$, while ensuring that the correlation between the unrelated query $\mathbf{q}$ and candidate $\mathbf{c}$ pairs is as low as possible.

In practice, for each query $\mathbf{q}$, we can collect its corresponding positive candidate $\mathbf{c}^+$ and a set of negative candidates $\mathcal{C}^-=\{\mathbf{c}_1^-, \dots, \mathbf{c}_K^- \}$, where $\mathcal{C}^-$ consists of both in-batch negatives and hard negatives, $K$ refers to the number of negative candidates. 
As aforementioned, both queries and candidates can encompass various modalities.
We employ contrastive learning with the InfoNCE loss \citep{oord2018infonce} to simultaneously minimize the distance between matched query-candidate pairs ($\mathbf{q}$, $\mathbf{c}^+$) while maximizing the distance between unmatched pairs ($\mathbf{q}$, $\mathbf{c}^-$):
\begin{equation}
\label{eq:infonce_loss}
\mathcal{L_\text{CL}}=-\log \frac{\exp ( \cos (\mathbf{q}, \mathbf{c}^+) / \tau )}{\exp ( \cos (\mathbf{q}, \mathbf{c}^+) / \tau ) + \sum_{\mathbf{c}_i^- \in \mathcal{C}^-}\exp ( \cos (\mathbf{q}, \mathbf{c}^-) / \tau )}
\end{equation}

where $\tau$ is the temperature hyper-parameter. 

\begin{figure}[t]
\centering
\includegraphics[width=\linewidth]{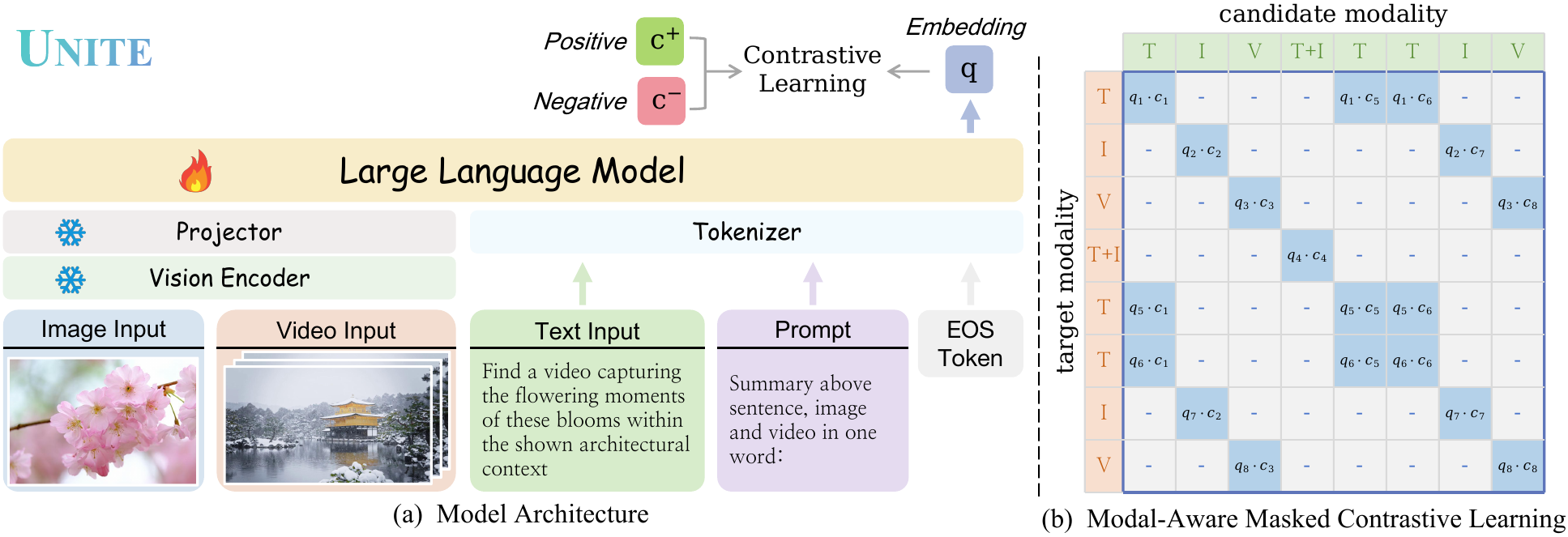}
\vspace{-1em}
\caption{Overview of \modelrgb: (a) Model architecture utilizing LMM as the backbone, supporting multimodal inputs (text, images, videos, and their combinations). (b) Similarity matrix after applying \modalmask, which enables focused contrastive learning by restricting comparisons to samples sharing the same target modality, thus reducing inter-modal interference.
}
\label{fig:model_arch}
\vspace{-1em}
\end{figure}

\subsection{Model Architecture}
\label{subsec:model_architecture}

Generally, LMMs are composed of three essential components: a large language model, a vision encoder, and a vision projector, as shown in Figure~\ref{fig:model_arch} (a). 
This architectural design empowers LMMs to handle text, images, videos, and their integrated forms with remarkable fluidity and efficiency. 
This native multimodal processing provides new promise for unified multimodal embeddings. 
Following recent methods~\cite{lin2024mmembed,jiang2024vlm2vec,zhang2024gme}, we extract target embeddings from the hidden state of the final token in the last layer.
Inspired by PromptEoL~\cite{jiang2024prompteol} and E5-V~\cite{jiang2024e5v}, we adapt the Explicit One word Limitation (EoL) method. 
Specifically, we use prompt template as follows:
\begin{center}
\texttt{<vision>\textbackslash{}n<text>\textbackslash{}nSummarize above <modalities> in one word:}
\end{center}
where \texttt{<vision>} and \texttt{<text>} are placeholders for visual content (\textit{i.e.}, images, videos) and textual sentences, respectively, and \texttt{<modalities>} specifies the input modality types. For instance, a video-text input would utilize the following prompt: ``\texttt{<video>\textbackslash{}n<text>\textbackslash{}nSummarize above video and sentence in one word:}''.

\subsection{Training Strategy}
\label{subsec:training_stategy}

We employ a two-stage training scheme: retrieval adaptation and instruction tuning. 
In the first stage, we adapt LMMs to general retrieval tasks, while we enhance the capabilities of LMMs for instruction-based complex retrieval scenarios in the second stage.

\textbf{Retrieval Adaptation.} LMMs have showcased remarkable prowess in multimodal understanding, adeptly handling and interpreting diverse forms of information. 
However, retrieval tasks remain uncharted territory for these models.
Thus, in the first stage, we focus on building robust fundamental retrieval capabilities by exposing LMMs to various information scenarios. It enables the LMMs to learn and adapt to the varying characteristics and requirements of different retrieval tasks.
For example, we utilize single-modal and multimodal retrieval data, including text-text, image-text and video-text pairs, in the experiments. 

\textbf{Instruction Tuning.} To enhance generalization across varied multimodal retrieval scenarios, we employ instruction tuning using comprehensive datasets like MMEB~\cite{jiang2024vlm2vec}, which covers 36 datasets across 4 multimodal tasks. 
This stage introduces complex fused-modal retrieval scenarios via instruction-guided samples. 
This goes beyond the basic single-modal and multimodal retrieval tasks, enabling a more sophisticated and nuanced understanding of retrieval tasks.

\subsection{Modal-Aware Masked Contrastive Learning}
\label{subsec:mamcl}

Traditional multimodal retrieval models typically employ standard InfoNCE loss~\cite{oord2018infonce} (\textit{i.e.}, Eq.~(\ref{eq:infonce_loss})) in the contrastive learning, it treats all negative pairs equally regardless of the modality compositions. 
However, this strategy overlooks the inherent distinctiveness of diverse modal combinations in retrieval tasks. 
For instance, embeddings derived solely from text and those from multimodal sources typically display substantial disparities in their distributions within the feature space.
When subjected to joint contrastive learning, 
the model struggles to balance the diverse information from different modalities, resulting in representations that fail to fully capture the semantic richness of each modality.
This leads to our hypothesis: 
\textit{Contrastive learning performed across instances featuring disparate target modalities has the potential to introduce noise and give rise to negative effects}.

To address this crucial problem, we propose \textit{Modal-Aware Masked Contrastive Learning} (\modalmask), introducing a modality-aware constraint to mitigate competitive relationships between various target modal instances. 
Given a batch of $N$ samples, we compute the similarity matrix $\mathbf{S} \in \mathbb{R}^{N\times (1+K)}$, where $1+K$ refers to the number of all positive and negative candidates,  $\mathbf{S}_{nk}$ represents the cosine similarity between query $\mathbf{q}_n$ and candidate $\mathbf{c}_k$:
%
\begin{equation}
\mathbf{S}_{nk} = \cos(\mathbf{q}_n, \mathbf{c}_{nk}) / \tau 
\end{equation}
%
where $n \in \{1, \dots, N\}$ and $k \in \{1, \dots, 1+K\}$, $\mathbf{c}_{nk}$ refers to the $k$-th candidate of the $n$-th query sample.
To incorporate modality awareness, we introduce a modality mask matrix $\pmb{\mathcal{M}} \in \{0,1\}^{N\times (1+K)}$, where $\pmb{\mathcal{M}}_{nk}$ indicates whether candidate $\mathbf{c}_{nk}$ shares the same target modality combination as the positive candidate $\mathbf{c}_{n}^+$ of query $\mathbf{q}_n$:
%
\begin{equation}
\pmb{\mathcal{M}}_{nk} = \mathbb{1}[\Phi(\mathbf{c}_{n}^+) = \Phi(\mathbf{c}_{nk})]
\end{equation}
%
where $\Phi(\cdot)$ refers to the operation of extracting the modality type of the embeddings. This extraction can be directly accomplished using the prior knowledge of the inputs.
In this manner, we ensure that each query only considers candidates with the same modality as its target candidate during the contrastive learning. 
Then, we update the masked similarity matrix $\tilde{\mathbf{S}}$ as follows:
%
\begin{equation}
\tilde{\mathbf{S}}_{nk} = \mathbf{S}_{nk} \cdot \pmb{\mathcal{M}}_{nk} + (-\infty) \cdot (1 - \pmb{\mathcal{M}}_{nk})
\end{equation}
%
A visualization of the similarity matrix $\tilde{\mathbf{S}}$ is shown in Figure \ref{fig:model_arch} (b). Finally, we expand the Eq.~(\ref{eq:infonce_loss}) to incorporate negatives, constantly ensuring that the model remains sensitive to different modalities:

\begin{equation}
\mathcal{L}_{\text{MAMCL}} = -\frac{1}{N} \sum_{n=1}^N \log \frac{\exp(\tilde{\mathbf{S}}_{np})}{\exp(\tilde{\mathbf{S}}_{np}) + \sum_{k=1, k\neq p}^{K+1} \exp(\tilde{\mathbf{S}}_{nk})}
\end{equation}
%
where $p$ refers to the index of the positive candidate. 

%% file: sections/4-experiments.tex
\section{Experiments}
\label{sec:experiments}

\input{tables/table-1-base}

\input{tables/table-2-3-mixed}

\paragraph{Experimental Setup} 
For the retrieval adaptation stage, we curate a diverse 7M-instance dataset spanning four categories: (1) text-text pairs from MSMARCO~\cite{Bajaj2016msmarco}, NLI~\cite{gao2021simcse}, NQ~\cite{kwiatkowski2019nq}, HotpotQA~\cite{yang2018hotpotqa}, Fever~\cite{thorne2018fever}, TriviaQA~\cite{joshi2017triviaqa} and SQuAD~\cite{rajpurkar2016squad}; (2) image-text pairs from CapsFusion~\cite{yu2024capsfusion}, LAION-Art~\cite{schuhmann2022laion} and MSCOCO~\cite{lin2014mscoco}; (3) video-text pairs from InternVid-10M-FLT~\cite{wang2023internvid}; (4) fine-grained video-caption pairs Tarsier2-Recap-585K~\cite{yuan2025tarsier2}. 
We name the model trained in this stage as \modelbase.
For the instruction tuning, we use the combination of MMEB~\cite{lin2024mmembed} and WebVid-CoVR~\cite{ventura2024covr2} as our training set. 
Thus, we name the model trained in this stage as \modelinstruct.
We will explain the underlying rationale for this data composition in Section~\ref{sec:analysis}.
Our specific training data composition and allocation are shown in Appendix~\ref{subapp:train-data-composition}.
We used Qwen2-VL \citep{wang2024qwen2vl} as the backbone of our model to conduct experiments on models with both 2B and 7B parameters. All evaluation datasets are described in Appendix \ref{subapp:evaluation-datasets}. All experimental results are reported in Recall@1 unless otherwise specified. We refer to Appendix~\ref{app:more-implementation} for more implementation details. 

\subsection{Main Results}
\label{subsec:main_results}

\paragraph{Fine-grained Retrieval.} Table \ref{tab:main_fine_video} demonstrates that our \modelbase achieves state-of-the-art performance and outperforms the existing methods with substantial margins, particularly on CaRe-General and CaRe-Spatial. 
This excellence stems from the incorporation of fine-grained video-caption pairs during retrieval adaptation, enhancing the feature representation capabilities of LMMs. 
While our 2B model outperforms all baselines on general and spatial retrieval tasks, its temporal retrieval performance remains moderate. After scaling the model size to 7B, we obtain significant improvements on general, spatial and temporal tasks. 
Notably, comparing our 2B and 7B models, we observe that model scaling achieves the most substantial relative improvements in temporal retrieval (\textit{e.g.}, 15.7\% and 10.8\%). 
It indicates a key insight: 
\textit{Models with larger sizes are likely to be more advantageous for retrieval tasks related to the temporal aspects of videos.}
Moreover, 
compared with the level that has been achieved in spatial tasks, there is still a great deal of room for improving temporal tasks. In addition, our \modelbase achieves leading performance in fine-grained image-text retrieval tasks, as shown in Table~\ref{tab:main_fine_image}. It indicates that our carefully designed data composition and allocation strategy indeed enables effective alignment across text, image, and video modalities.

\input{tables/table-4-mmeb}

\paragraph{Instruction-based Retrieval.} 

As shown in Table~\ref{tab:main_covr}, \modelinstruct 2B  substantially outperforms existing models on the WebVid-CoVR-Test~\citep{ventura2024covr}. Similarly, increasing the scale of model size to 7B can significantly boost the improvement margins. 
In Table~\ref{tab:main_mmeb}, the evaluation on the MMEB benchmark, encompassing 36 datasets across four meta-tasks, demonstrates superior performance of \modelinstruct against various existing models at different parameter scales. 
For example, \model surpasses both larger-scale models (\textit{e.g.}, mmE5 11B~\cite{chen2025mme5} and IDMR 26B~\cite{liu2025idmr}) and models trained with more extensive datasets (\textit{e.g.}, MMRet~\cite{zhou2024megapairs} with 26M image-text retrieval samples). 
These compelling results across text, image, and video retrieval scenarios can be attributed to our evolving training strategy and MAMCL strategy (See details in Section~\ref{subsec:ablation}).

\textbf{Coarse-grained Cross-Modal Retrieval.} 
We also assess our the performance of \model on coarse-grained retrieval tasks. As demonstrated in Table~\ref{tab:main_coarse_image} and Table~\ref{tab:main_coarse_video}, our models achieve competitive results on coarse-grained cross-modal retrieval tasks, including text-image and text-video scenarios. It proves that our data composition strategy effectively balances performance across different modalities on various retrieval tasks.

\input{tables/table-5-6-zeroshot}

\subsection{Ablation Study}
\label{subsec:ablation}

\input{tables/table-7-ablation}

\paragraph{Modal-Aware Masked Contrastive Learning.} 
We evaluate the effectiveness of Modal-Aware Masked Contrastive Learning (MAMCL) through comprehensive ablation studies across diverse instruction-based retrieval scenarios.
As shown in Table~\ref{tab:abl_mamcl} (3$\to$4), we observe that there exists significant performance improvements on MMEB after integrating the MMEB training set. However, the performance degrades on WebVid-CoVR. It verifies our hypothesis that cross-modal interfere might occur among samples with distinct target modalities. 
Results in Table~\ref{tab:abl_mamcl} (1$\to$2, 4$\to$5, 6$\to$7) reveal that MAMCL successfully mitigates these inter-modal effects. Specifically, MAMCL yields substantial improvements in in-distribution (IND) scenarios, validating its effectiveness in 
scenarios where test samples align with the training distribution. 
While exhibiting minor fluctuations on out-of-distribution (OOD) datasets, the performance indicates that its generalization capabilities remain well.

\begin{wraptable}{ht}{0.45\textwidth}
\centering
\vspace{-0.5em}
\caption{Ablation study on hard negative. We report the MMEB overall and WebVid-CoVR R@1 scores.}
\label{tab:abl_hard_negative}
\vspace{+0.5em}
\resizebox{0.45\textwidth}{!}{
\begin{tabular}{l|ll|l}
\toprule
\textbf{Setting} & \textbf{MMEB} & \textbf{CoVR} & \textbf{Avg} \\ 
\midrule
\modelinstruct 2B & 63.3 & 69.1 & 66.2 \\
w/o hard-negative & 62.4 \down{0.9} & 68.0 \down{1.1} & 65.2 \down{1.0} \\
\bottomrule
\end{tabular}%
}
\end{wraptable}

\textbf{Hard Negatives.} Previous methods~\cite{robinson2020contrastive,xiong2020approximate,radenovic2023filtering} have evidenced the importance of hard negatives for retrieval tasks. 
We conduct an ablation experiment on our \modelinstruct 2B model. 
As shown in Table~\ref{tab:abl_hard_negative}, model without the hard negative leads to performance degradation, which underscores the critical role it plays in the training of effective retrieval models.

\section{Analysis}
\label{sec:analysis}

In this section, we present a systematic investigation of two critical aspects: (1) the impact of various training data on different retrieval tasks, and (2) efficient training strategies for enhancing the fine-grained retrieval capabilities of LMMs.

\paragraph{Training Data Composition}
Understanding the optimal composition of training data for text-image-video retrieval scenarios remains an open research problem that warrants systematic investigation. 
We conduct comprehensive experiments utilizing Text-Text (TT), Text-Image (TI), and Text-Video (TV) datasets, both independently and in various combined scenarios (See details in Appendix~\ref{subapp:more-implementation-analysis-data}). Finally, the evaluation covers coarse-grained, fine-grained, and instruction-based retrieval tasks. 

\textit{Video-text pairs prove to be superior training data for general cross-modal retrieval.} 
The first striking finding is that the TV-only training pattern consistently outperforms all other configurations across the entire spectrum of cross-modal retrieval tasks, as shown in the ``Coarse'' and ``Fine'' results in Table~\ref{tab:main_analysis_modal} (See more details in Table~\ref{tab:app_analysis_modal_general} in Appendix~\ref{app:more-results-analysis}). 
Notably, in image-text retrieval tasks, training solely with TV data outperforms training that uses only TI data.
This result represents a novel discovery, as it contradicts the established findings in conventional image-text research, challenging existing assumptions about the optimal data sources for such retrieval tasks.
This unexpected outcome underscores the importance of reevaluating traditional data selection strategies and opens new avenues for exploring how different data types interact with model training in cross-modal retrieval scenarios.

\begin{table}[!htbp]
\centering
\caption{Results of various training data composition on different retrieval tasks in retrieval adaptation stage. To ensure fairness, the total data size for all configurations is 600K. All scores of cross-modal retrieval are the average of R@1 in zero-shot setting. The tested datasets include (1) coarse-grained image-text datasets (Flickr30K, MSCOCO), video-text datasets (MSR-VTT, MSVD); (2) fine-grained image-text dataset (DOCCI), video-text dataset (CaReBench); and (3) instruction-based datasets (MMEB, WebVid-CoVR).}
\label{tab:main_analysis_modal}%
\resizebox{\textwidth}{!}{
\begin{tabular}{ccc|cccc|cccc|cccc}
\toprule
\multicolumn{3}{c|}{\textbf{Setting}} & \multicolumn{2}{c}{\textbf{Coarse I-T}} & \multicolumn{2}{c|}{\textbf{Coarse V-T}} & \multicolumn{2}{c}{\textbf{Fine I-T}} & \multicolumn{2}{c|}{\textbf{Fine V-T}} & \multicolumn{3}{c}{\textbf{MMEB}} & \textbf{CoVR} \\
\cmidrule(lr){1-3}\cmidrule(lr){4-5}\cmidrule(lr){6-7}\cmidrule(lr){8-9}\cmidrule(lr){10-11}\cmidrule(lr){12-14}\cmidrule(lr){15-15}
TT & TI & TV & T$\to$I & I$\to$T & T$\to$V & V$\to$T & T$\to$I & I$\to$T & T$\to$V & V$\to$T & IND & OOD & Overall & R@1 \\
\midrule
\cmark & & & 53.5 & 64.1 & 37.8 & 45.8 & 69.1 & 65.4 & 45.5 & 52.4 & 61.9 & 58.5 & 60.4 & 64.4 \\
& \cmark & & 55.4 & 68.9 & 40.5 & 51.0 & 75.8 & 71.8 & 57.9 & 62.9 & 62.6 & 58.3 & 60.7 & \best{66.5} \\
& & \cmark & \best{60.2} & \best{73.8} & \best{44.3} & \best{56.0} & \best{79.8} & \best{74.9} & \best{65.8} & \best{68.7} & 62.6 & 59.1 & 61.1 & 65.6 \\
\midrule
\cmark & \cmark & & 55.6 & 67.6 & 41.6 & 50.8 & 74.8 & 70.3 & 56.8 & 58.8 & \best{63.8} & \best{59.9} & \best{62.1} & 65.4 \\
\cmark & & \cmark & 56.9 & 65.1 & 43.4 & 54.6 & 76.3 & 70.1 & 62.3 & 64.1 & 62.1 & 57.7 & 60.2 & 65.6 \\
& \cmark & \cmark & \second{58.3} & \second{71.8} & \second{43.7} & \second{55.9} & \second{77.8} & \second{73.0} & \second{65.7} & \second{68.4} & \second{63.0} & \second{59.2} & \second{61.3} & \second{65.8} \\
\midrule
\cmark & \cmark & \cmark & 58.1 & 67.8 & 43.0 & 54.6 & 76.5 & 70.9 & 61.3 & 61.2 & 62.7 & 59.0 & 61.0  & 64.8 \\
\bottomrule
\end{tabular}%
}
\end{table}

\textit{Text-text and text-image pairs are essential for instruction-following tasks.} To further investigate the impact of data composition on broader retrieval tasks, we conducted comprehensive experiments on instrcution-based retrieval tasks. 
TT+TI training overall outperforms other combinations on instruction-based retrieval tasks, including TV-only configuration that excel in general cross-modal retrieval tasks, as shown in the ``MMEB'' and ``CoVR'' results in Table~\ref{tab:main_analysis_modal} (See more details in Table \ref{tab:app_analysis_modal_instruct} in Appendix~\ref{app:more-results-analysis}). 
This observation can be attributed to two key factors: (1) Text-text pairs enhance linguistic understanding and logical reasoning capabilities, establishing a solid and comprehensive foundation for interpreting complex retrieval instructions.
(2) Text-image pairs provide precise multimodal alignment information, empowering more focused semantic connections compared to video content. 
These factors enable the model to capture detailed vision-language correspondences that are crucial for instruction following. 

\paragraph{Effectively Utilizing Fine-Grained Video-Caption Data}
Recent advances in video LMMs have produced powerful captioning models (\textit{e.g.}, Tarsier \citep{wang2024tarsier}, AuroraCap \citep{chai2024auroracap}) and fine-grained datasets like LLaVA-Video-178K \citep{zhang2024llavavideo}. 
While CaRe has shown that fine-tuning LMMs with these video-caption pairs prior to retrieval adaptation can significantly boost the performance of fine-grained video retrieval, 
a notable limitation exists in its retrieval adaptation phase, which depends solely on text-text pairs. 
This raises two critical questions: (1) Does the \textit{fine-grained alignment} continue to yield effective results when video-text pairs are integrated into the retrieval adaptation process?
and (2) How can we optimally utilize fine-grained video-text pairs to achieve the greatest possible enhancements in performance?

\begin{table}[!htbp]
\centering
\caption{Ablation study on fine-grained alignment stage (\textbf{Align}) across distinct retrieval-adaptation data (\textbf{Retrieval}). The reported scores are Recall@1 of zero-shot results. The composition and volume of retrieval-adaptation data used for ID 1-6 are consistent with those in Table~\ref{tab:main_analysis_modal} in Appendix~\ref{app:more-results-analysis}.}
\label{tab:abl_alignment}%
\resizebox{\textwidth}{!}{
\begin{tabular}{ccc|llllllllll|l}
\toprule
\multirow{3}[3]{*}{\textbf{ID}} & \multicolumn{2}{c|}{\multirow{2}[2]{*}{\textbf{Setting}}} & \multicolumn{4}{c}{\textbf{Coarse-grained Video-Text Retrieval}} & \multicolumn{6}{c|}{\textbf{Fine-grained Video-Text Retrieval}} & \multirow{3}[3]{*}{\textbf{Avg}} \\
\cmidrule(lr){4-7}\cmidrule(lr){8-13}
& & & \multicolumn{2}{c}{MSR-VTT} & \multicolumn{2}{c}{MSVD} & \multicolumn{2}{c}{CaRe-General} & \multicolumn{2}{c}{CaRe-Spatial} & \multicolumn{2}{c|}{CaRe-Temporal} \\
\cmidrule(lr){2-3}\cmidrule(lr){4-5}\cmidrule(lr){6-7}\cmidrule(lr){8-9}\cmidrule(lr){10-11}\cmidrule(lr){12-13}
& Align & Retrieval & T$\to$V & V$\to$T & T$\to$V & V$\to$T & T$\to$V & V$\to$T & T$\to$V & V$\to$T & T$\to$V & V$\to$T \\
\midrule
1 & \xmark & TT & 32.9 & 31.5 & 42.7 & 60.1 & 45.5 & 52.4 & 47.7 & 51.3 & 33.1 & 37.7 & 43.5 \\
2 & \cmark & TT & 33.5 \up{0.6} & 30.8 \down{0.7} & 42.5 \down{0.2} & 60.1 \tie{} & 50.1 \up{4.6} & 51.5 \down{0.9} & 51.3 \up{3.6} & 51.4 \up{0.1} & 34.8 \up{1.7} & 37.7 \tie{} & 44.4 \up{0.9} \\
\midrule
3 & \xmark & TV & \best{41.5} & \best{41.7} & 47.0 & \best{70.3} & 65.8 & 68.7 & 68.3 & 67.3 & 42.0 & \best{47.1} & 56.0 \\
4 & \cmark & TV & 39.6 \down{1.9} & 40.4 \down{1.3} & \best{47.4} \up{0.4} & \best{70.3} \tie{} & 68.5 \up{2.7} & 68.1 \down{0.6} & 67.6 \down{0.7} & 65.9 \down{1.4} & 41.6 \down{0.4} & 45.3 \down{1.8} & 55.5 \down{0.5} \\
\midrule
5 & \xmark & TT+TI+TV & 40.0 & 39.2 & 46.0 & 69.9 & 61.3 & 61.2 & 63.0 & 59.6 & 40.5 & 42.7 & 52.3 \\
6 & \cmark & TT+TI+TV & 38.1 \down{1.9} & 39.0 \down{0.2} & 45.6 \down{0.4} & 69.1 \down{0.8} & 59.1 \down{2.2} & 60.2 \down{1.0} & 62.1 \down{0.9} & 58.8 \down{0.8} & 39.7 \down{0.8} & 43.1 \up{0.4} & 51.5 \down{0.8} \\
\midrule
7 & \xmark & Fine TV & 34.8 & 35.6 & 42.3 & 69.6 & \best{81.2} & \best{79.9} & \best{81.0} & \best{80.5} & \best{42.9} & 46.0 & \best{59.4} \\
\midrule
8 & \xmark & TV + Fine TV & 39.9 & 39.1 & 40.6 & 69.6 & 79.2 & 79.0 & 79.4 & 76.7 & 41.8 & 45.7 & \best{59.4} \\
\bottomrule    
\end{tabular}
}
\end{table}

To solve these two questions, we conduct extensive experiments across diverse settings (See details in Appendix~\ref{subapp:more-implementation-analysis-finecaption}).
Our research endeavors focus on evaluating the effectiveness of \textit{fine-grained alignment}. This is achieved by performing next token prediction fine-tuning on 500K fine-grained video-caption instances sourced from Tarsier2-Recap-585K~\cite{yuan2025tarsier2} with various configurations. 
Initial experiments validate the findings in CaRe~\cite{xu2025carebench}: when text-text (TT) data is employed for retrieval adaptation, fine-grained alignment significantly boosts performance, especially in fine-grained video-text retrieval tasks (See Table~\ref{tab:abl_alignment}, 1$\to$2). 
However, when text-video (TV) pairs are involved in retrieval adaptation, fine-grained alignment surprisingly leads to performance degradation (See Table~\ref{tab:abl_alignment}, 3$\to$4, 5$\to$6). Our results reveal several key insights:
\begin{itemize}[leftmargin=*, nolistsep, noitemsep]
    \item During the retrieval adaptation process, leveraging TV pairs yields far more significant performance improvements than those obtained through fine-grained alignment.
    \item The exclusive employment of fine-grained video-text pairs during the retrieval adaptation process leads to a substantial enhancement in CaReBench performance. However, it causes a severe degradation in the model's coarse-grained retrieval capabilities (See Table~\ref{tab:abl_alignment}, 7). 
    \item By incorporating fine-grained TV pairs into general TV data (See Table \ref{tab:abl_alignment}, 8), we can achieves a balanced performance. It enables the model to obtain competitive results in both coarse- and fine-grained video-text retrieval tasks. 
\end{itemize}

Thus, these observations lead to a crucial insight: \textit{During the retrieval adaptation, the direct incorporation of fine-grained video-caption pairs has been shown to be far more effective than the implementing a isolated fine-grained alignment stage}.

%% file: tables/table-1-base.tex
\begin{table}[t]
\centering
\caption{Zero-shot performance of fine-grained video-text retrieval on CaReBench~\cite{xu2025carebench}. LLaVA-NV refers to LLaVA-NeXT-Video~\cite{zhang2024llavanextvideo}. $\dagger$ indicates that these models have been equipped with feature representation capabilities through contrastive learning. \textit{Rel.}$\Delta$ represents the relative performance improvement of \modelbase~7B compared to \modelbase~2B. 
Results in \best{bold} and \second{underline} denote the best and second-best performances.}
\resizebox{\textwidth}{!}{
\begin{tabular}{l|cccc|cccc|cccc}
\toprule
\multirow{3}[3]{*}{\textbf{Model}} & \multicolumn{4}{c|}{\textbf{CaRe-General}} & \multicolumn{4}{c|}{\textbf{CaRe-Spatial}} & \multicolumn{4}{c}{\textbf{CaRe-Temporal}} \\
\cmidrule(lr){2-5}\cmidrule(lr){6-9}\cmidrule(lr){10-13}
& \multicolumn{2}{c}{T$\to$V} & \multicolumn{2}{c|}{V$\to$T} & \multicolumn{2}{c}{T$\to$V} & \multicolumn{2}{c|}{V$\to$T} & \multicolumn{2}{c}{T$\to$V} & \multicolumn{2}{c}{V$\to$T} \\
\cmidrule(lr){2-3}\cmidrule(lr){4-5}\cmidrule(lr){6-7}\cmidrule(lr){8-9}\cmidrule(lr){10-11}\cmidrule(lr){12-13}
& R@1 & R@5 & R@1 & R@5 & R@1 & R@5 & R@1 & R@5 & R@1 & R@5 & R@1 & R@5 \\
\midrule
\rowcolor{gray!10}\multicolumn{13}{l}{\textit{\textbf{CLIP-based Models}}}\\
\midrule
CLIP L/14~\cite{radford2021clip} & 51.2 & 83.4 & 54.7 & 86.9 & 49.0 & 81.9 & 55.4 & 85.6 & 33.5 & 70.3 & 39.7 & 76.2 \\
LanguageBind~\cite{zhu2023languagebind} & 64.3 & 91.0 & 59.5 & 88.0 & 64.7 & 90.8 & 61.0 & 87.2 & 39.8 & 77.3 & 42.2 & 77.6 \\
Long-CLIP L/14~\cite{zhang2024longclip} & 62.7 & 88.8 & 60.3 & 88.8 & 65.6 & 90.9 & 61.0 & 88.3 & 33.2 & 68.8 & 34.5 & 71.9 \\
\text{InternVideo2}$_\text{stage2}$ 1B~\cite{wang2024internvideo2} & 72.5 & 93.7 & 69.5 & 94.6 & 72.4 & 94.2 & 62.7 & 90.5 & 46.0 & 80.8 & 46.6 & 82.5 \\
\midrule
\rowcolor{gray!10}\multicolumn{13}{l}{\textit{\textbf{LMM-based Models}}}\\
\midrule
LLaVA-NV 7B$^\dagger$~\cite{zhang2024llavanextvideo} & 66.9 & 89.4 & 62.7 & 89.2 & 68.0 & 92.0 & 65.0 & 90.0 & 43.3 & 76.9 & 40.1 & 75.4 \\
MiniCPM-V 2.6$^\dagger$~\cite{yao2024minicpmv} & 71.0 & 92.2 & 69.3 & 92.8 & 71.7 & 93.6 & 67.6 & 92.3 & 50.5 & 82.9 & 46.1 & 80.9 \\
InternVL2 8B$^\dagger$~\cite{chen2024internvl1.5} & 72.1 & 92.6 & 73.6 & 93.4 & 76.1 & 94.1 & 74.3 & 94.5 & 48.1 & 76.8 & 47.6 & 78.2 \\
Tarsier 7B$^\dagger$~\cite{wang2024tarsier} & 71.0 & 93.8 & 70.6 & 94.2 & 70.2 & 94.0 & 67.4 & 93.5 & 50.1 & 84.1 & 50.0 & 84.7 \\
Qwen2-VL 7B$^\dagger$~\cite{wang2024qwen2vl} & 76.6 & 95.3 & 77.4 & 95.6 & 78.2 & 95.5 & 75.4 & 95.0 & \second{51.9} & \second{84.8} & 52.7 & 85.4 \\
CaRe 7B~\cite{xu2025carebench} & 77.0 & \second{95.6} & 79.0 & \second{96.8} & 76.8 & \second{96.3} & \second{78.1} & \second{95.8} & 50.7 & \best{85.3} & \second{53.4} & \second{86.3} \\
\midrule
\modelbase 2B & \second{78.1} & 95.5 & \second{80.8} & 96.4 & \second{79.6} & 95.4 & 78.0 & 95.4 & 45.3 & 77.6 & 50.0 & 83.6 \\
\modelbase 7B & \best{86.0} & \best{96.8} & \best{86.9} & \best{98.3} & \best{86.5} & \best{96.9} & \best{84.8} & \best{98.0} & \best{52.4} & 82.5 & \best{55.4} & \best{86.5} \\
\cmidrule(lr){1-1}\cmidrule(lr){2-5}\cmidrule(lr){6-9}\cmidrule(lr){10-13}
\rowcolor{LightCyan}\textit{Rel.}$\Delta$ & \relativeimprove{10.1}& \relativeimprove{1.4} & \relativeimprove{7.5} & \relativeimprove{2.0} & \relativeimprove{8.7} & \relativeimprove{1.6} & \relativeimprove{8.7} & \relativeimprove{1.6} & \relativeimprove{15.7} & \relativeimprove{6.3} & \relativeimprove{10.8} & \relativeimprove{3.5} \\
\bottomrule
\end{tabular}%
}

\label{tab:main_fine_video}%
\end{table}

%% file: tables/table-2-3-mixed.tex
\begin{table}[t]
\centering
\begin{minipage}{0.5\textwidth}
\scriptsize
\setlength\tabcolsep{2.2pt}
\caption{Results of fine-grained image-text retrieval on ShareGPT4V, Urban1K, and DOCCI.}
\resizebox{0.9\textwidth}{!}{
\begin{tabular}{l|cccccc}
\toprule
\multirow{2}[2]{*}{\textbf{Model}} & \multicolumn{2}{c}{\textbf{ShareGPT4V}} & \multicolumn{2}{c}{\textbf{Urban1K}} & \multicolumn{2}{c}{\textbf{DOCCI}}   \\
\cmidrule(lr){2-3}\cmidrule(lr){4-5}\cmidrule(lr){6-7}
& T$\to$I & I$\to$T & T$\to$I & I$\to$T & T$\to$I & I$\to$T \\
\midrule
\rowcolor{gray!10}\multicolumn{7}{l}{\textit{\textbf{CLIP-based Models}}}\\
CLIP L/14 \citep{radford2021clip} & 83.6 & 84.2 & 55.6 & 68.3 & 65.8 & 63.1 \\
OpenCLIP L/14 \citep{cherti2023openclip} & 81.8 & 84.0 & 47.0 & 47.0 & -    & -    \\
Long-CLIP L/14 \citep{zhang2024longclip} & \best{97.3} & \best{97.2} & 86.1 & 82.7 & 78.6 & 66.5 \\
EVA-CLIP 8B  \citep{sun2023evaclip} & 93.1 & 91.2 & 80.4 & 77.8 & -    & -    \\
FineLIP \citep{asokan2025finelip} & -    & -    & 94.1 & 93.2 & \second{86.0} & \second{84.5} \\
\midrule
\rowcolor{gray!10}\multicolumn{7}{l}{\textit{\textbf{LMM-based Models}}}\\
MATE \citep{jang2024mate} & -    & -    & -    & -    & 84.6 & 76.6 \\
E5-V 7B \citep{jiang2024e5v} & 85.1 & 82.1 & 88.9 & 83.2 & -    & -    \\
VLM2Vec 7B \citep{jiang2024vlm2vec} & 90.7 & 85.8 & 90.8 & 84.7 & -    & -    \\
UniME 7B \citep{gu2025unime} & \second{93.9} & \best{97.2} & \second{95.2} & \best{95.9} & -    & -  \\ 
\midrule
\modelbase 2B & 89.7 & 86.7 & 91.5 & 89.2 & 81.4 & 72.3 \\
\modelbase 7B & 93.3 & \second{93.2} & \best{95.5} & \second{95.6} & \best{87.2} & \best{85.8} \\
\bottomrule
\end{tabular}%
}
\label{tab:main_fine_image}
\end{minipage}
\hspace{0.01\textwidth}
\begin{minipage}{0.45\textwidth}
\scriptsize
\caption{Results on the WebVid-CoVR. We take the best variant of baselines in the table.}
\setlength\tabcolsep{2pt}
\resizebox{0.9\textwidth}{!}{
\begin{tabular}{l|cccc}
\toprule
\multirow{2}[2]{*}{\textbf{Model}} & \multicolumn{4}{c}{\textbf{WebVid-CoVR-Test (TV$\to$V)}} \\
\cmidrule(lr){2-5}
& R@1 & R@5 & R@10 & R@50 \\
\midrule
\rowcolor{gray!10}\multicolumn{5}{l}{\textit{\textbf{Zero-shot Setting}}}\\
LanguageBind~\cite{zhu2023languagebind} & 43.2 & 66.3 & 75.2 & -    \\
CoVR~\cite{ventura2024covr} & 45.5 & 70.5 & 79.5 & 93.3 \\
CoVR-2~\cite{ventura2024covr2} & 45.7 & 71.7 & 81.3 & 94.8 \\
TFR-CVR~\cite{hummel2024egocvr} & 51.7 & 75.3 & 80.7 & -    \\
\midrule
\rowcolor{gray!10}\multicolumn{5}{l}{\textit{\textbf{Finetuning Setting}}}\\
CoVR~\cite{ventura2024covr} & 53.1 & 79.9 & 86.9 & 97.7 \\
CoVR-2~\cite{ventura2024covr2} & 59.8 & 83.8 & 91.3 & 98.2 \\
ECDE~\cite{thawakar2024covrecde} & 60.1 & 84.3 & 91.3 & 98.7 \\
\midrule
\modelinstruct 2B & \second{69.1} & \second{88.4} & \second{93.2} & \second{99.1} \\
\modelinstruct 7B & \best{72.5} & \best{90.8} & \best{95.3} & \best{99.5} \\
\bottomrule
\end{tabular}%
}
\label{tab:main_covr}
\end{minipage}
\end{table}

%% file: tables/table-4-mmeb.tex
\begin{table}[t]
\centering
\caption{Results on the MMEB benchmark~\cite{jiang2024vlm2vec}. We average the scores in each meta-task. We compare numerous recent works, across diverse model scales. ``IND'' refers to the in-distribution dataset, and ``OOD'' denotes the out-of-distribution dataset.}
\label{tab:main_mmeb}
\resizebox{\textwidth}{!}{
\begin{tabular}{lcccccccc}
\toprule
\multirow{2}[2]{*}{\textbf{Model}} & \multirow{2}[2]{*}{\textbf{\#Parameters}} & \multicolumn{4}{c}{\textbf{Per Meta-Task Score}} & \multicolumn{3}{c}{\textbf{Average Score}} \\ 
\cmidrule(lr){3-6} \cmidrule(lr){7-9}
& & Classification & VQA & Retrieval & Grounding & IND & OOD & Overall \\ \midrule
\#Datasets $\rightarrow$ & & 10 & 10 & 12 & 4 & 20 & 16 & 36 \\
\midrule
\rowcolor{gray!10}\multicolumn{9}{l}{\textit{\textbf{Zero-shot Setting}}}\\
\midrule
Magiclens (ViT-L/14)~\cite{zhang2024magiclens} & 0.4B
& 38.8 & 8.3  & 35.4 & 26.0 & 31.0 & 23.7 & 27.8 \\
CLIP (ViT-L/14)~\cite{radford2021clip} & 0.4B
& 42.8 & 9.1  & 53.0 & 51.8 & 37.1 & 38.7 & 37.8 \\
OpenCLIP (ViT-L/14)~\cite{cherti2023openclip} & 0.4B
& 47.8 & 10.9 & 52.3 & 53.3 & 39.3 & 40.2 & 39.7 \\
BLIP2 (ViT-L/14)~\cite{li2023blip2} & 1.2B
& 27.0 & 4.2  & 33.9 & 47.0 & 25.3 & 25.1 & 25.2 \\
\midrule
\rowcolor{gray!10}\multicolumn{9}{l}{\textit{\textbf{Finetuning Setting}}}\\
\midrule
CLIP (ViT-L/14)~\cite{radford2021clip} & 0.4B
& 55.2 & 19.7 & 53.2 & 62.2 & 47.6 & 42.8 & 45.4 \\
OpenCLIP (ViT-L/14)~\cite{cherti2023openclip} & 0.4B
& 56.0 & 21.9 & 55.4 & 64.1 & 50.5 & 43.1 & 47.2 \\
E5-V (LLaVA-1.6)~\cite{jiang2024e5v} & 7B
& 39.7 & 10.8 & 39.4 & 60.2 & 34.2 & 33.4 & 33.9 \\
MMRet-MLLM (LLaVA-1.6)~\cite{zhou2024megapairs} & 7B
& 56.0 & 57.4 & 69.9 & 83.6 & 68.0 & 59.1 & 64.1 \\
VLM2Vec (Qwen2-VL, high-res)~\cite{jiang2024vlm2vec} & 7B
& 62.6 & 57.8 & 69.9 & 81.7 & 72.2 & 57.8 & 65.8 \\
UniME (LLaVA-1.6)~\cite{gu2025unime}& 7B
& 60.6 & 52.9 & 67.9 & 85.1 & 68.4 & 57.9 & 66.6 \\
CAFe (LLaVA-OV)~\cite{yu2025cafe}& 7B
& 65.2 & \best{65.6} & 70.0 & \best{91.2} & \best{75.8} & 62.4 &  \second{69.8} \\
mmE5 (Llama-3.2-Vision)~\cite{chen2025mme5} & 11B
& \second{67.6} & 62.8 & 70.9 & \second{89.7} & 72.3 & \best{66.7} & \second{69.8} \\
IDMR (InternVL2.5)~\cite{liu2025idmr} & 26B
& 66.3 & 61.9 & \second{71.1} & 88.6 & 73.4 & 63.9 & 69.2 \\
\midrule
\modelinstruct 2B & 2B
& 63.2 & 55.9 & 65.4 & 75.6 & 65.8 & 60.1 & 63.3 \\
\modelinstruct 7B & 7B
& \best{68.3} & \second{65.1} & \best{71.6} & 84.8 & \second{73.6} & \second{66.3} & \best{70.3} \\
\bottomrule
\end{tabular}%
}
\end{table}

%% file: tables/table-5-6-zeroshot.tex
\begin{table}[h]
\centering
\begin{minipage}[t]{0.4\textwidth}
\scriptsize
\setlength\tabcolsep{1.6pt}
\caption{Zero-shot coarse-grained image-text retrieval results on Flickr30K~\cite{plummer2015flickr30k}.}
\label{tab:main_coarse_image}
\resizebox{0.98\textwidth}{!}{
\begin{tabular}{l|cccc}
\toprule
\multirow{2}[3]{*}{\textbf{Model}}  & \multicolumn{2}{c}{T$\to$I} & \multicolumn{2}{c}{I$\to$T} \\
\cmidrule(lr){2-3}\cmidrule(lr){4-5}
& R@1 & R@5 & R@1 & R@5 \\
\midrule
OpenCLIP-L~\cite{cherti2023openclip} & 75.0 & 92.5 & 88.7 & 98.4 \\
MagicLens-L~\cite{zhang2024magiclens} & 79.7 & 95.0 & 89.6 & 98.7 \\
\midrule
CAFe 7B~\cite{yu2025cafe}          & 75.3 & 92.6 & 87.5 & 98.2 \\
VLM2Vec 7B~\cite{jiang2024vlm2vec} & 80.3 & 95.0 & 94.6 & 99.5 \\
LamRA-Ret 7B~\cite{liu2024lamra}   & 82.8 & -    & 92.7 & -    \\
UniME 7B~\cite{gu2025unime}        & 81.9 & -    & 93.4 & -    \\
\midrule
\modelbase 2B & 80.9 & 95.3 & 89.6 & 98.4 \\
\modelbase 7B & 86.1 & 96.9 & 94.4 & 99.5 \\
\bottomrule
\end{tabular}%
}
\end{minipage}
\hspace{0.01\textwidth}
\begin{minipage}[t]{0.56\textwidth}
\centering
\scriptsize
\caption{Zero-shot coarse-grained video-text retrieval results on MSR-VTT~\cite{xu2016msrvtt}, MSVD~\cite{chen2011msvd}, and DiDeMo~\cite{anne2017didemo}.}
\label{tab:main_coarse_video}
\setlength\tabcolsep{1.6pt}
\resizebox{0.95\textwidth}{!}{
\begin{tabular}{l|cccccc}
\toprule
\multirow{2}[2]{*}{\textbf{Model}} & \multicolumn{2}{c}{\textbf{MSR-VTT}} & \multicolumn{2}{c}{\textbf{MSVD}} & \multicolumn{2}{c}{\textbf{DiDeMo}} \\
\cmidrule(lr){2-3}\cmidrule(lr){4-5}\cmidrule(lr){6-7}
& T$\to$V & V$\to$T & T$\to$V & V$\to$T & T$\to$V & V$\to$T \\
\midrule
InternVideo~\cite{wang2022internvideo} & 40.7 & 39.6 & 43.4 & 67.6 & 31.5 & 33.5 \\
LanguageBind~\cite{zhu2023languagebind} & 42.1 & 65.9 & 40.1 & 65.4 & 35.6 & 35.6 \\
ViCLIP~\cite{wang2023internvid} & 42.4 & 41.3 & 49.1 & 75.1 & 18.4 & 27.9 \\
\midrule
VLM2Vec 7B~\cite{jiang2024vlm2vec} & 43.5 & -    & 49.5 & -    & -    & -    \\
LamRA 7B~\cite{liu2024lamra}    & 44.7 & -    & 52.4 & -    & -    & -    \\
CaRe 7B~\cite{xu2025carebench}  & 43.9 & 41.7 & 52.6 & 74.6 & 41.4 & 39.1 \\
\midrule
\modelbase 2B & 43.8 & 41.7 & 50.0 & 73.1 & 37.9 & 37.5 \\
\modelbase 7B & 46.5 & 45.2 & 50.4 & 76.1 & 43.5 & 40.3 \\
\bottomrule
\end{tabular}%
}
\end{minipage}
\end{table}

%% file: tables/table-7-ablation.tex
\begin{table}[!htbp]
\centering
\caption{Ablation study of our proposed MAMCL. We show results of various settings on the MMEB and WebVid-CoVR test sets. \gray{Gray} indicates without fine-tuning on the corresponding training set. \textbf{Avg} refers to the average of the overall score on MMEB and the R@1 score on WebVid-CoVR.}
\label{tab:abl_mamcl}
\resizebox{0.95\textwidth}{!}{
\begin{tabular}{cccc|lll|lll|l}
\toprule
\multirow{2}[2]{*}{\textbf{ID}} & \multicolumn{3}{c|}{\textbf{Setting}} & \multicolumn{3}{c|}{\textbf{MMEB}} & \multicolumn{3}{c|}{\textbf{WebVid-CoVR}} & \multirow{2}[2]{*}{\textbf{Avg}} \\ 
\cmidrule(lr){2-4}\cmidrule(lr){5-7}\cmidrule(lr){8-10}
& MMEB & CoVR & MAMCL & IND & OOD & Overall & R@1 & R@5 & R@10 \\
\midrule
\rowcolor{gray!10}\multicolumn{11}{c}{\textit{\textbf{2B parameters}}}\\
\midrule
1 & \cmark & & & 64.4 & 60.1 & 62.5 & \gray{49.6} & \gray{74.5} & \gray{82.1} & - \\
2 & \cmark & & \cmark & 65.5 \up{1.1} & 60.2 \up{0.1} & 63.1 \up{0.6} & \gray{48.0} & \gray{73.4} & \gray{80.9} & - \\
\midrule
3 & & \cmark & & \gray{36.1} & \gray{34.9} & \gray{35.6} & 69.2 & 89.4 & 93.6 & - \\
\midrule
4 & \cmark & \cmark & & 64.8 & 60.3 & 62.8 & 67.4 & 88.0 & 92.6 & 65.1 \\
5 & \cmark & \cmark & \cmark & 65.8 \up{1.0} & 60.1 \down{0.2} & 63.3 \up{0.5} & 69.1 \up{1.7} & 88.4 \up{0.4} & 93.2 \up{0.6} & 66.2 \up{1.1} \\
\midrule
\rowcolor{gray!10}\multicolumn{11}{c}{\textit{\textbf{7B parameters}}}\\
\midrule
6 & \cmark & \cmark & & 73.3 & 65.8 & 70.0 & 71.4 & 91.1 & 94.7 & 70.7 \\
7 & \cmark & \cmark & \cmark & 73.6 \up{0.3} & 66.3 \up{0.5} & 70.3 \up{0.3} & 72.5 \up{1.1} & 90.8 \down{0.3} & 95.3 \up{0.6} & 71.4 \up{0.7} \\
\bottomrule
\end{tabular}%
}
\end{table}

%% file: sections/5-conclusion.tex
\section{Conclusion}
\label{sec:conclusion}

In this work, we introduce \modelrgb, a universal multimodal embedding framework that enables the seamlessly integration of text, image, and video modalities. 
Through systematic analysis of how training with varied data compositions affects the final retrieval performance, we observe novel insights that have not received limited exploration in image-text and video-text retrieval scenarios. 
Based on these insights, we propose an data composition and allocation strategy, and introduce MAMCL to mitigate inter-instance competition while maintaining balance the representation learning across text, images, and videos. 
Extensive experimental results demonstrate that \modelrgb achieves state-of-the-art results on 40+ tasks spanning coarse-grained, fine-grained, and instruction-based retrieval scenarios. 
We believe that our work advances the development of unified multimodal retrieval and provides valuable insights for future research.

%% file: sections/appendix.tex
\appendix

\begin{figure}[h]
   \centering
   \includegraphics[width=0.9\linewidth]{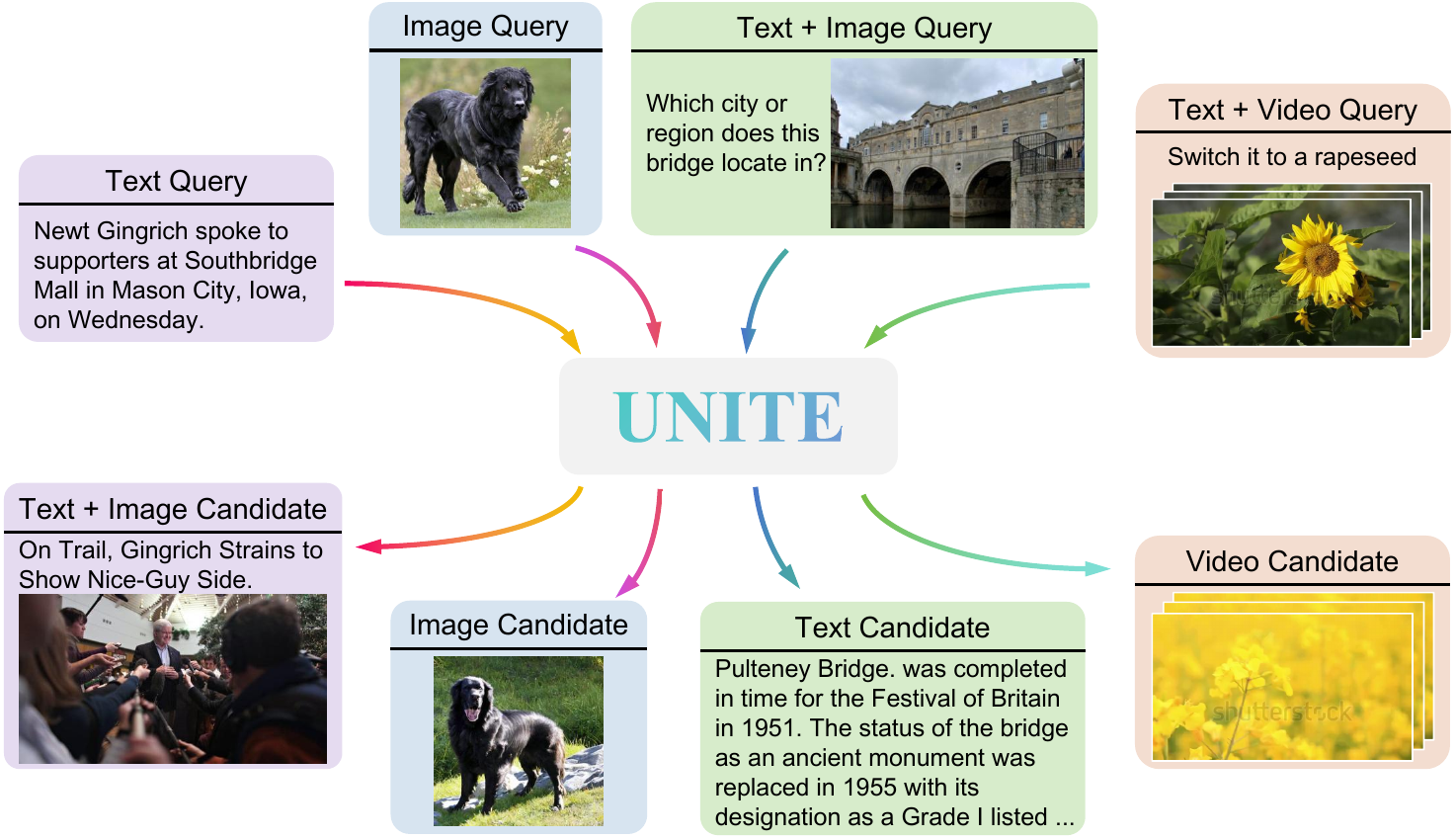}
   \caption{We develop a universal multimodal embedder \modelrgb, allowing for a unified representation of arbitrary multimodal contents.}
   \label{fig:overall}
\end{figure}

\begin{figure}[htbp]
\centering
\includegraphics[width=0.7\textwidth]{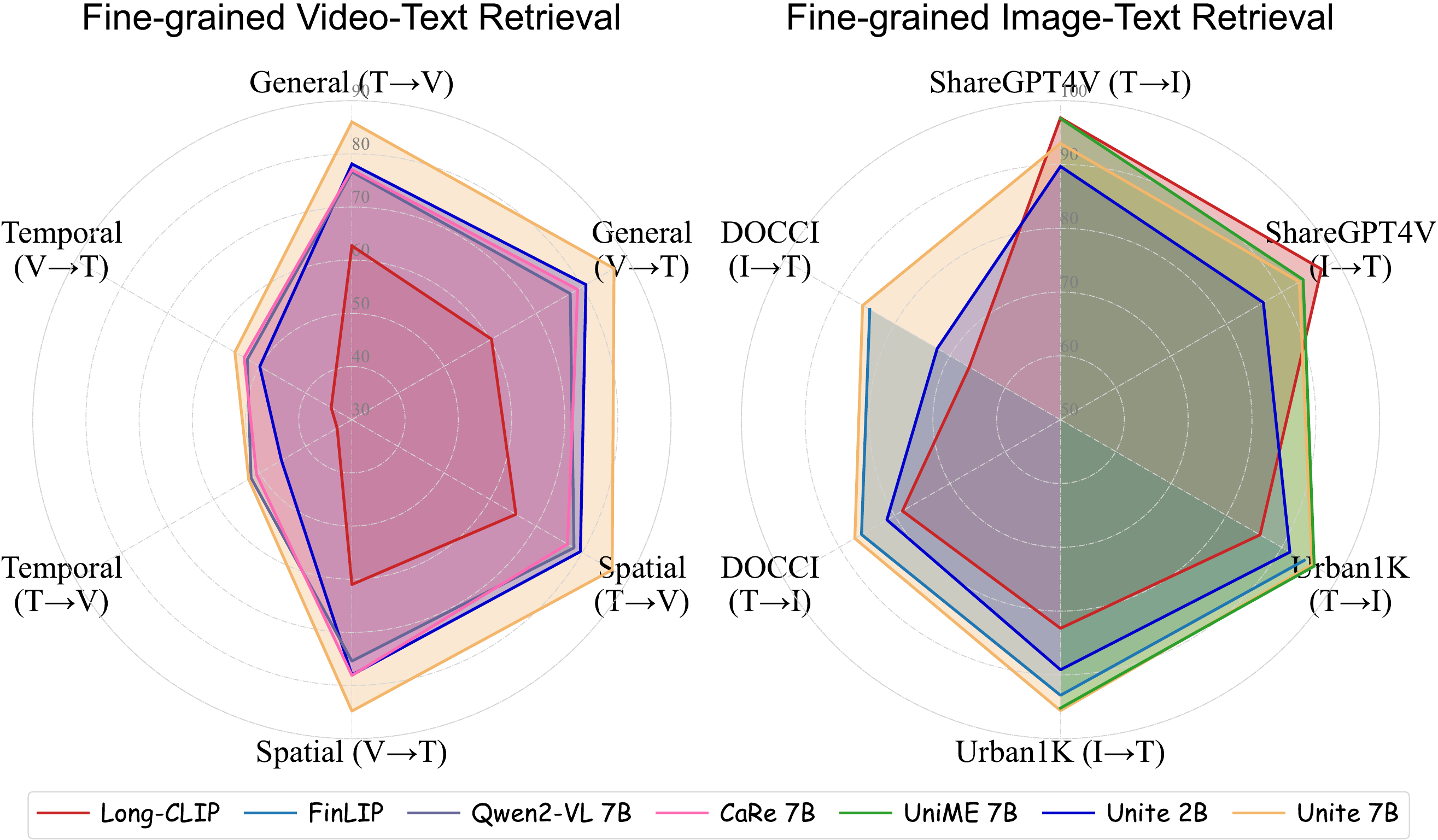}
\caption{Performance comparison on fine-grained video-text benchmark (CaReBench~\cite{xu2025carebench}) and image-text benchmarks (ShareGPT4V~\cite{chen2024sharegpt4v}, Urban1K~\cite{zhang2024longclip}, DOCCI~\cite{onoe2024docci}). Our \modelrgb achieves the overall optimal performance.}
\label{fig:radar_fine}
\end{figure}

\section{Data}
\label{app:data}

\subsection{Training Data Composition}
\label{subapp:train-data-composition}

Our thoroughly analysis across coarse-grained, fine-grained, and instruction-based retrieval tasks reveals that text-video pairs are particularly effective for cross-modal retrieval, while text-text and text-image pairs are essential for instruction-based retrieval (Section~\ref{sec:analysis}). Based on these findings, we carefully curate the retrieval adaptation data to achieve a balanced performance across different modalities. The detailed data composition is illustrated in Figure~\ref{fig:pie_adaptation_data}.

For instruction tuning, surpassing existing works that solely rely on supervised learning on MMEB~\cite{jiang2024vlm2vec}, we combine 20 instruction-based image-text retrieval datasets from MMEB with a subset of WebVid-CoVR~\cite{ventura2024covr} to optimize performance across text, image, and video modalities. The detailed data composition is illustrated in Figure~\ref{fig:pie_instrcut_data}.

\input{tables/app-data-adaptation}

\input{tables/app-data-instruct}

\subsection{Evaluation Datasets}
\label{subapp:evaluation-datasets}

We provide brief descriptions of all evaluation benchmarks, and their statistics are shown in Table \ref{tab:eval_statistics}.

\input{tables/app-data-eval}
\subsubsection{Fine-grained Retrieval Datasets}
\label{subapp:evaluation-datasets-fine}

\textbf{CaReBench}~\cite{xu2025carebench} consists of 1,000 video-caption pairs with hierarchical annotations covering overall summary, static objects, dynamic actions, and miscellaneous aspects. Its distinctive feature lies in the manually annotated spatial and temporal information. We evaluate fine-grained video-text retrieval through three tasks: General, Spatial, and Temporal.

\textbf{ShareGPT4V}~\cite{chen2024sharegpt4v} features comprehensive image-text pairs with rich descriptions of world knowledge, object properties, spatial relationships, and aesthetic elements. The dataset comprises 100K GPT4-Vision generated captions and 1.2M model-expanded captions, averaging 942 characters in length. Following Long-CLIP~\cite{zhang2024longclip}, we utilize 1K instances for testing.

\textbf{Urban1K}~\cite{zhang2024longclip} is derived from Urban-200 and expanded to 1,000 urban scene image-text pairs. Each image is paired with a GPT-4V generated caption (averaging 101 words) detailing object types, colors, and spatial relationships. The dataset challenges models with visually similar urban scenes, requiring fine-grained cross-modal understanding.

\textbf{DOCCI}~\cite{onoe2024docci} (Descriptions of Connected and Contrasting Images) contains 15,000 images with human-annotated descriptions (averaging 136 words). Curated by a single researcher, it evaluates spatial relations, counting, text rendering, and world knowledge comprehension. The dataset features contrast sets with subtle variations in object arrangements. We evaluate using the official 5K test split.

\subsubsection{Instruction-based Retrieval Datasets}
\label{subapp:evaluation-datasets-instruct}

\textbf{MMEB}~\cite{jiang2024vlm2vec} is a comprehensive multimodal embedding benchmark comprising 36 datasets across four meta-tasks: classification, visual question answering, retrieval, and visual grounding. The benchmark is strategically divided into 20 in-distribution training datasets and 16 out-of-distribution evaluation datasets. All tasks are formulated as ranking problems with 1,000 candidates, where models process instruction-guided queries (text, images, or both) to select correct targets.

The meta-tasks are structured as follows:
\begin{itemize}
    \item \textit{Classification}: Queries combine instructions with images (and optional text) to predict class labels.
    \item \textit{Visual Question Answering}: Queries include instructions, images, and questions, with answers as targets.
    \item \textit{Information Retrieval}: Both queries and targets can be multimodal combinations.
    \item \textit{Visual Grounding}: Queries pair instructions with full images to locate specific objects, using cropped regions as candidates.
\end{itemize}

MMEB spans diverse domains (common, news, Wikipedia, web, fashion) and supports various instruction types from object recognition to retrieval tasks. Performance is evaluated using Precision@1, measuring the accuracy of selecting the correct candidate from 1,000 options. The benchmark's comprehensive design makes it an ideal testbed for universal multimodal embeddings.

\textbf{WebVid-CoVR-Test}~\cite{ventura2024covr} is a manually verified benchmark for Composed Video Retrieval (CoVR), containing 2,435 high-quality video-text-video triplets. Each triplet consists of a query video, a modification text describing desired changes, and a target video. The benchmark is derived from WebVid10M, specifically excluding WebVid2M content to ensure evaluation integrity. The dataset features diverse modification texts (averaging 4.8 words) and videos (averaging 16.8 seconds), covering a wide range of content variations.

\input{tables/app-mmeb-statistics}
\subsubsection{Coarse-grained Retrieval Datasets}
\label{subapp:evaluation-datasets-coarse}

\textbf{Flickr30K}~\cite{plummer2015flickr30k} consists of 31K images collected from Flickr, with each image paired with five human-annotated captions describing its content. The dataset covers diverse everyday scenarios and human activities in natural settings. Following common practice~\cite{karpathy2015deep}, we use the standard split with 1,000 images for testing.

\textbf{MSRVTT}~\cite{xu2016msrvtt} comprises 10K video clips paired with 200K captions, covering diverse topics including human activities, sports, and natural landscapes. For text-to-video retrieval evaluation, we adopt the standard \texttt{1K-A split} following prior works.

\textbf{MSVD}~\cite{chen2011msvd} features 1,970 videos, with approximately 40 captions annotated per video. 

\textbf{DiDeMo}~\cite{anne2017didemo} features 10K long-form Flickr videos, where each video is annotated with four temporally ordered sentences. Following previous works, we concatenate these sentences for paragraph-to-video retrieval evaluation using the official split.

\section{More Implementation Details}
\label{app:more-implementation}

We simply incorporate LoRA~\cite{hu2022lora} module into LMM, with a rank of 8, and gradually unlock its retrieval capabilities. We employ a temperature of 0.03 for contrastive learning, with learning rates of 1e-4 and 2e-5 for retrieval adaptation and instruction tuning phases, respectively. Following Qwen2-VL~\cite{wang2024qwen2vl}, we process images with dynamic resolution, constraining the number of image tokens to (256, 1,280). For video processing, we sample frames at 1fps with a 12-frame cap, limiting each frame's maximum number of tokens to 98. Our experiments are conducted on 64 NVIDIA A100 GPUs for training and 8 NVIDIA H100 GPUs for evaluation. The following experiments adopt these default settings unless specified otherwise.

\subsection{Stage1: Retrieval Adaptation}
\label{subapp:more-implementation-stage1}

During retrieval adaptation, we follow the standard InfoNCE loss (\textit{i.e.}, Eq.~(\ref{eq:infonce_loss})) and employ a bidirectional contrastive learning strategy to maximize the utilization of training data. Taking image-text pairs as an example, we simultaneously optimize both text-to-image and image-to-text retrieval directions. The final loss function is computed as the average of these bidirectional retrieval processes:

\begin{equation}
\mathcal{L}_{\text{bi}}=\frac{1}{2} \left[ 
-\frac{1}{N} \sum_{n=1}^N \log \frac{\exp \left( \cos (\mathbf{q}_n, \mathbf{c}_n^+) / \tau \right)}{\sum_{j=1}^N \exp \left( \cos\left(\mathbf{q}_n, \mathbf{c}_j\right) / \tau \right)} 
-\frac{1}{N} \sum_{n=1}^N \log \frac{\exp \left( \cos (\mathbf{c}_n^+, \mathbf{q}_n) / \tau \right)}{\sum_{j=1}^N \exp \left( \cos\left(\mathbf{c}_n^+, \mathbf{q}_j\right) / \tau \right)}
\right]
\end{equation}

The training data distribution and hyperparameters are shown in Figure \ref{fig:pie_adaptation_data} and Table \ref{tab:training-parameters-base}, respectively.

\subsection{Stage2: Instruction Tuning}
\label{subapp:more-implementation-stage2}

During instruction tuning, we observe that jointly training on instruction-based retrieval data from different modalities (image-text and video-text) may lead to performance fluctuations, with potential degradation in certain domains. To address this challenge and maintain consistent performance across modalities, we propose MAMCL (Modality-Aware Masked Contrastive Learning) to balance the learning dynamics between different modal data. This approach helps optimize the overall multimodal retrieval capabilities while preserving domain-specific performance. The detailed data distribution and training parameters are presented in Figure~\ref{fig:pie_instrcut_data} and Table~\ref{tab:training-parameters-instruct}.

\input{tables/app-training-hyperparameters}

\subsection{Analysis: Training Data Composition}
\label{subapp:more-implementation-analysis-data}

We conduct comprehensive experiments utilizing Text-Text (TT), Text-Image (TI), and Text-Video (TV) datasets, both independently and in various combined scenarios.
Specifically, we collect TT data from MSMARCO~\cite{Bajaj2016msmarco} and NLI~\cite{gao2021simcse}, TI data from CapsFusion~\cite{yu2024capsfusion}, and TV data from InternVid~\cite{wang2023internvid}. 
To ensure fair comparisons, we maintain a consistent total dataset size of 600K instances across all configurations, with mixed-pattern settings employing equal distribution of different constituent types (\textit{e.g.}, 200K per dataset in TT+TI+TV). 

The experimental pipeline consists of two phases. In the first phase, we finetune Qwen2-VL-2B~\cite{wang2024qwen2vl} using 600K instances across seven distinct configurations, following the hyperparameters specified in Table~\ref{tab:training-parameters-base}. The second phase involves instruction tuning, where we independently finetune the retrieval-adapted model using the complete MMEB~\cite{jiang2024vlm2vec} training set and 500K instances from WebVid-CoVR~\cite{ventura2024covr}, respectively, adhering to the hyperparameters outlined in Table~\ref{tab:training-parameters-instruct}.

To thoroughly investigate the impact of various data compositions on retrieval performance, we evaluate across three distinct retrieval scenarios, including (1) coarse-grained image-text datasets (Flickr30K~\cite{plummer2015flickr30k}, MSCOCO~\cite{lin2014mscoco}), video-text datasets (MSR-VTT~\cite{xu2016msrvtt}, MSVD~\cite{chen2011msvd}); (2) fine-grained image-text dataset (DOCCI~\cite{onoe2024docci}), video-text dataset (CaReBench~\cite{xu2025carebench}); and (3) instruction-based datasets (MMEB~\cite{jiang2024vlm2vec}, WebVid-CoVR~\cite{ventura2024covr}). The raw results are shown in Table \ref{tab:app_analysis_modal_general} and Table \ref{tab:app_analysis_modal_instruct}.

\subsection{Analysis: Effectively Utilizing Fine-Grained Video-Caption Data}
\label{subapp:more-implementation-analysis-finecaption}

To investigate efficient strategies for fine-grained video-caption training, we conduct preliminary experiments using Tarsier2-Recap-585K~\cite{yuan2025tarsier2}. Extending the exploration scope of CaRe's~\cite{xu2025carebench} work, we examine the effectiveness of fine-grained alignment across broader data compositions (TT, TV, and TT+TI+TV). This expanded investigation aims to understand the generalizability of \textit{fine-grained alignment} (\textit{i.e.}, fine-tuning LMM through next token prediction using fine-grained video-caption pairs) in diverse multimodal scenarios.

The experimental configurations in Table~\ref{tab:abl_alignment} (IDs 1-6) utilize 500K instances from Tarsier2-Recap for fine-grained alignment, maintaining consistent retrieval adaptation data as described in Section~\ref{subapp:more-implementation-analysis-data}. Configuration ID 7 omits the fine-grained alignment, instead directing all 500K fine-grained video-caption pairs to retrieval adaptation. Similarly, ID 8 bypasses fine-grained alignment and employs a balanced training set with 600K instances, comprising 300K instances each from InternVid and Tarsier2-Recap.

For fine-grained alignment, we leverage \texttt{ms-swift}\footnote{\url{https://github.com/modelscope/ms-swift}} to train LoRA modules of Qwen2-VL-2B, configured with a rank of 16 and an alpha of 32. We set the batch size to 128 and the learning rate to 1e-4. For video items, we sample frames at 1fps with a 16-frame cap, limiting each frame's maximum number of tokens to 192.

Notably, unified contrastive learning demonstrates superiority not only in performance but also in computational efficiency. Our empirical analysis reveals significant training speed advantages: processing 500K instances from Tarsier2-Recap requires 4.2 hours for fine-grained alignment and merely 0.9 hours for retrieval adaptation, highlighting the method's computational efficiency while maintaining superior performance.

\section{More Experimental Results}
\label{app:more-results}

\subsection{Detailed Main Results}
\label{app:more-results-main}

Detailed experimental results on CaReBench, MMEB, and coarse-grained video-text retrieval benchmarks are provided in Tables \ref{tab:app_fine_video}, \ref{tab:app_mmeb_full}, and \ref{tab:app_coarse_video}, in correspondence with Tables \ref{tab:main_fine_video}, \ref{tab:main_mmeb}, and \ref{tab:main_coarse_video}. Table \ref{tab:app_mmeb_per_task} provides the results of MMEB specifically for the 36 tasks. Figure \ref{fig:radar_fine} shows the visualization of the fine-grained retrieval results.

\input{tables/app-care-full}

\input{tables/app-mmeb-full}

\input{tables/app-coarse-video-text-full}

\subsection{Detailed Results for Analysis}
\label{app:more-results-analysis}

Tables \ref{tab:app_analysis_modal_general}, \ref{tab:app_analysis_modal_instruct} show the raw results of Table \ref{tab:main_analysis_modal}.

\input{sections/limitations}

\begin{table}[!htbp]
\centering
\caption{Impact study of training data composition on general retrieval in the retrieval adaptation stage. We report zero-shot cross-modal retrieval results on coarse-grained cross-modal datasets (Flickr30K \citep{plummer2015flickr30k}, MSCOCO \citep{lin2014mscoco}, MSR-VTT \citep{xu2016msrvtt}, MSVD \citep{chen2011msvd}) and fine-grained cross-modal datasets (DOCCI \citep{onoe2024docci}, CaReBench-General\citep{xu2025carebench}) with Recall@1.}
\label{tab:app_analysis_modal_general}%
\resizebox{\textwidth}{!}{
\begin{tabular}{ccc|cccccccccccc|c}
\toprule
\multicolumn{3}{c|}{\multirow{2}[2]{*}{\textbf{Setting}}} & \multicolumn{4}{c}{\textbf{Coarse Image-Text}} & \multicolumn{4}{c}{\textbf{Coarse Video-Text}} & \multicolumn{2}{c}{\textbf{Fine Image-Text}} & \multicolumn{2}{c|}{\textbf{Fine Video-Text}} & \multirow{3}[3]{*}{\textbf{Avg}} \\
\cmidrule(lr){4-7}\cmidrule(lr){8-11}\cmidrule(lr){12-13}\cmidrule(lr){14-15}
& & & \multicolumn{2}{c}{Flickr30K} & \multicolumn{2}{c}{COCO} & \multicolumn{2}{c}{MSR-VTT} & \multicolumn{2}{c}{MSVD} & \multicolumn{2}{c}{DOCCI} & \multicolumn{2}{c|}{CaRe-General}\\
\cmidrule(lr){1-3}\cmidrule(lr){4-5}\cmidrule(lr){6-7}\cmidrule(lr){8-9}\cmidrule(lr){10-11}\cmidrule(lr){12-13}\cmidrule(lr){14-15}
TT & TI & TV & T$\to$I & I$\to$T & T$\to$I & I$\to$T& T$\to$V & V$\to$T & T$\to$V & V$\to$T  & T$\to$I & I$\to$T & T$\to$V & V$\to$T \\
\midrule
\cmark & & & 66.1 & 78.2 & 40.8 & 49.9 & 32.9 & 31.5 & 42.7 & 60.1 & 69.1 & 65.4 & 45.5 & 52.4 & 52.9 \\
& \cmark & & 68.1 & 79.5 & 42.6 & 58.2 & 37.2 & 35.6 & 43.7 & 66.4 & 75.8 & 71.8 & 57.9 & 62.9 & 58.3 \\
& & \cmark & \best{73.4} & \best{86.9} & \best{47.0} & \best{60.6} & \best{41.5} & \best{41.7} & \best{47.0} & \best{70.3} & \best{79.8} & \best{74.9} & \best{65.8} & \best{68.7} & \best{63.1} \\
\midrule
\cmark & \cmark & & 67.7 & 78.9 & 43.5 & 56.3 & 38.5 & 35.2 & 44.6 & 66.3 & 74.8 & 70.3 & 56.8 & 58.8 & 57.6 \\
\cmark & & \cmark & 70.9 & 82.0 & 42.8 & 48.2 & 40.4 & 39.5 & \second{46.4} & 69.7 & 76.3 & 70.1 & 62.3 & 64.1 & 59.4 \\
& \cmark & \cmark & 71.6 & \second{84.2} & \second{45.0} & \second{59.3} & \second{41.3} & 41.4 & 46.0 & \best{70.3} & \second{77.8} & \second{73.0} & \second{65.7} & \second{68.4} & \second{62.0} \\
\midrule
\cmark & \cmark & \cmark & \second{71.7} & 82.0 & 44.4 & 53.6 & 40.0 & 39.2 & 46.0 & \second{69.9} & 76.5 & 70.9 & 61.3 & 61.2 & 59.7 \\
\bottomrule
\end{tabular}%
}
\end{table}

\begin{table}[!htbp]
\centering
\caption{Results of diverse training data composition on instruction-based retrieval in the retrieval adaptation stage.}
\label{tab:app_analysis_modal_instruct}%
\resizebox{\textwidth}{!}{
\begin{tabular}{ccc|ccccccc|ccc}
\toprule
\multicolumn{3}{c|}{\textbf{Setting}} & \multicolumn{7}{c|}{\textbf{MMEB}} & \multicolumn{3}{c}{\textbf{WebVid-CoVR}} \\
\cmidrule(lr){1-3}\cmidrule(lr){4-10} \cmidrule(lr){11-13}
TT & TI & TV & Classification & VQA & Retrieval & Grounding & IND & OOD & Overall & R@1 & R@5 & R@10 \\
\midrule
\cmark & & & 60.9 & 53.8 & 61.1 & 73.7 & 61.9 & 58.5 & 60.4 & 64.4 & 86.3 & 92.1 \\
& \cmark & & 60.5 & 55.2 & 61.2 & 73.3 & 62.6 & 58.3 & 60.7 & \best{66.5} & 87.2 & 92.3 \\
& & \cmark & 60.9 & 55.0 & \second{61.9} & 74.1 & 62.6 & 59.1 & 61.1 & 65.6 & \best{87.8} & 92.4 \\
\midrule
\cmark & \cmark & & \best{61.9} & \second{55.4} & \best{62.8} & \best{77.1} & \best{63.8} & \best{59.9} & \best{62.1} & 65.4 & 87.0 & 92.1 \\
\cmark & & \cmark & 59.7 & 54.4 & 60.7 & 74.2 & 62.1 & 57.7 & 60.2 & 65.6 & \second{87.4} & \second{92.6} \\
& \cmark & \cmark & \second{61.5} & \best{55.5} & 61.6 & 74.3 & \second{63.0} & \second{59.2} & \second{61.3} & \second{65.8} & \second{87.4} & \second{92.6} \\
\midrule
\cmark & \cmark & \cmark & 61.2 & 54.7 & 61.8 & \second{74.6} & 62.7 & 59.0 & 61.0 & 64.8 & 86.8 & \best{92.7} \\
\bottomrule
\end{tabular}%
}
\end{table}

\input{tables/app-mmeb-per-task}

%% file: tables/app-data-adaptation.tex
\begin{figure}[htp]
\centering
\begin{minipage}{.3\textwidth}
\includegraphics[height=125pt]{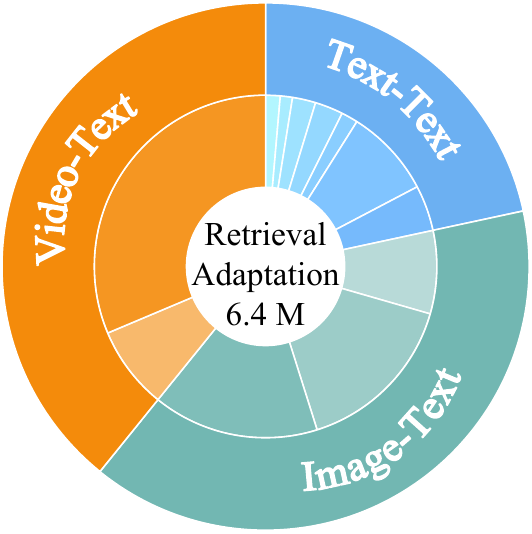}
\end{minipage}%
\begin{minipage}{.7\textwidth}
\centering
\renewcommand{\arraystretch}{1.1}
\setlength\tabcolsep{3pt}
\fontsize{6pt}{9pt}\selectfont
\begin{tabular}{lll}
\makecell[l]{\parbox[c]{2.2cm}{\cellcolor[RGB]{108,176,242} \textcolor{white}{Text-Text Pairs (21.6\%)}}} & 
\tikz[baseline=0.05em] \fill [color={rgb,255: red,118; green,186; blue,252}] (0,0) rectangle (0.75em,0.75em); NLI \citep{gao2021simcse} (275.6 K) &
\tikz[baseline=0.05em] \fill [color={rgb,255: red,138; green,206; blue,255}] (0,0) rectangle (0.75em,0.75em); MSMARCO \citep{Bajaj2016msmarco} (532.8 K) \\
\tikz[baseline=0.05em] \fill [color={rgb,255: red,158; green,226; blue,255}] (0,0) rectangle (0.75em,0.75em); NQ \citep{kwiatkowski2019nq} (100.2 K) &
\tikz[baseline=0.05em] \fill [color={rgb,255: red,178; green,246; blue,255}] (0,0) rectangle (0.75em,0.75em); HotpotQA \citep{yang2018hotpotqa} (170.0 K) & 
\tikz[baseline=0.05em] \fill [color={rgb,255: red,198; green,255; blue,255}] (0,0) rectangle (0.75em,0.75em); Fever \citep{thorne2018fever}(140.1 K) \\
\tikz[baseline=0.05em] \fill [color={rgb,255: red,218; green,255; blue,255}] (0,0) rectangle (0.75em,0.75em); TriviaQA (73.3 K) & 
\tikz[baseline=0.05em] \fill [color={rgb,255: red,228; green,255; blue,255}] (0,0) rectangle (0.75em,0.75em); SQuAD~\cite{rajpurkar2016squad} (87.6 K) \\
\hline

\makecell[l]{\parbox[c]{2.2cm}{\cellcolor[RGB]{60,206,173} \textcolor{white}{Image-Text Pairs (39.2\%)}}} & 
\tikz[baseline=0.05em] \fill [color={rgb,255: red,126; green,202; blue,169}] (0,0) rectangle (0.75em,0.75em); CapsFusion \citep{yu2024capsfusion} (1.0 M) &
\tikz[baseline=0.05em] \fill [color={rgb,255: red,189; green,236; blue,216}] (0,0) rectangle (0.75em,0.75em); LAION-Art \citep{schuhmann2022laion} (1.0 M) \\
\tikz[baseline=0.05em] \fill [color={rgb,255: red,239; green,255; blue,249}] (0,0) rectangle (0.75em,0.75em); MSCOCO \citep{lin2014mscoco} (500 K) \\
\hline

\makecell[l]{\parbox[c]{2.2cm}{\cellcolor[RGB]{255,140,0} \textcolor{white}{Video-Text Pairs (36.1\%)}}} & 
\tikz[baseline=0.05em] \fill [color={rgb,255: red,255; green,150; blue,0}] (0,0) rectangle (0.75em,0.75em); InternVid-FLT~\cite{wang2023internvid} (2.0 M) &
\tikz[baseline=0.05em] \fill [color={rgb,255: red,255; green,180; blue,0}] (0,0) rectangle (0.75em,0.75em); Tarsier2-Recap (500.0 K) \\
\end{tabular}
\end{minipage}
\caption{\textbf{Retrieval Adaptation 6.4M.} Left: Data Distribution within Each Category. The outer circle shows the distribution of all data categories and the inner circle shows the distribution of data subsets. Right: The detailed quantities of datasets.}
\label{fig:pie_adaptation_data}
\end{figure}

%% file: tables/app-data-instruct.tex
\begin{figure}[htp]
\centering
\begin{minipage}{.3\textwidth}
\includegraphics[height=125pt]{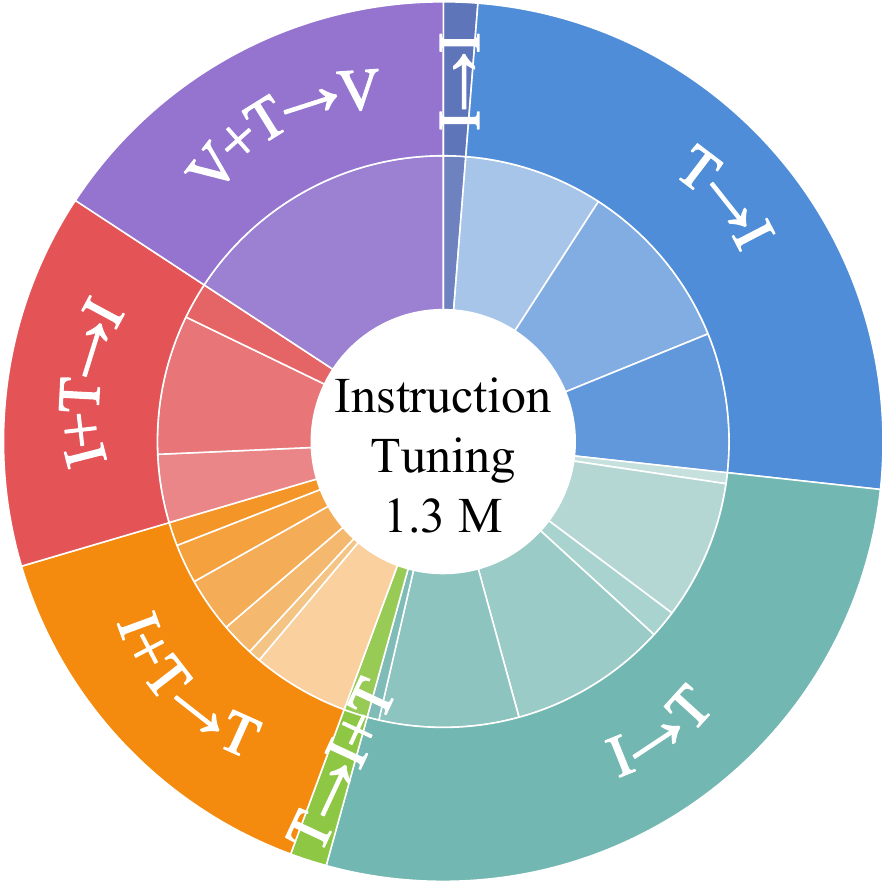}
\end{minipage}%
\begin{minipage}{.7\textwidth}
\centering
\renewcommand{\arraystretch}{1.1}
\setlength\tabcolsep{3pt}
\fontsize{6pt}{9pt}\selectfont
\begin{tabular}{lll}
\makecell[l]{\parbox[c]{2.5cm}{\cellcolor[RGB]{94,117,186} \textcolor{white}{Image$\to$Image (1.3\%)}}} & 
\tikz[baseline=0.05em] \fill [color={rgb,255: red,110; green,130; blue,192}] (0,0) rectangle (0.75em,0.75em); NIGHTS~\cite{fu2023nights} (15.9 K) \\
\hline

\makecell[l]{\parbox[c]{2.5cm}{\cellcolor[RGB]{80,141,216} \textcolor{white}{Text$\to$Image (25.5\%)}}} & 
\tikz[baseline=0.05em] \fill [color={rgb,255: red,97; green,151; blue,219}] (0,0) rectangle (0.75em,0.75em);  MSCOCO~\cite{lin2014mscoco} (100.0 K) &
\tikz[baseline=0.05em] \fill [color={rgb,255: red,130; green,173; blue,226}] (0,0) rectangle (0.75em,0.75em);  VisDial~\cite{das2017visdial} (123.3 K) \\
\tikz[baseline=0.05em] \fill [color={rgb,255: red,166; green,197; blue,233}] (0,0) rectangle (0.75em,0.75em);  VisualNews~\cite{liu2020visualnews} (99.9 K) \\

\hline
\makecell[l]{\parbox[c]{2.5cm}{\cellcolor[RGB]{114,183,177} \textcolor{white}{Image$\to$Text (27.6\%)}}} & 
\tikz[baseline=0.05em] \fill [color={rgb,255: red,127; green,188; blue,183}] (0,0) rectangle (0.75em,0.75em);  HatefulMemes~\cite{kiela2020hatefulmemes} (8.5 K) &
\tikz[baseline=0.05em] \fill [color={rgb,255: red,141; green,196; blue,192}] (0,0) rectangle (0.75em,0.75em);  ImageNet-1K~\cite{deng2009imagenet} (100.0 K) \\
\tikz[baseline=0.05em] \fill [color={rgb,255: red,155; green,203; blue,199}] (0,0) rectangle (0.75em,0.75em);  MSCOCO~\cite{lin2014mscoco} (113.3 K) &
\tikz[baseline=0.05em] \fill [color={rgb,255: red,169; green,211; blue,207}] (0,0) rectangle (0.75em,0.75em);  SUN397~\cite{xiao2010sun397} (19.9 K) &
\tikz[baseline=0.05em] \fill [color={rgb,255: red,181; green,215; blue,212}] (0,0) rectangle (0.75em,0.75em);  VisualNews~\cite{liu2020visualnews} (100.0 K) \\
\tikz[baseline=0.05em] \fill [color={rgb,255: red,198; green,224; blue,222}] (0,0) rectangle (0.75em,0.75em);  VOC2007~\cite{everingham2015voc2007} (7.9 K) \\
\hline
\makecell[l]{\parbox[c]{2.5cm}{\cellcolor[RGB]{141,199,68}  \textcolor{white}{Text$\to$Image+Text (1.4\%)}}} & 
\tikz[baseline=0.05em] \fill [color={rgb,255: red,151; green,203; blue,86}] (0,0) rectangle (0.75em,0.75em);  WebQA~\cite{chang2022webqa} (17.2 K) \\
\hline

\makecell[l]{\parbox[c]{2.5cm}{\cellcolor[RGB]{244,139,14} \textcolor{white}{Image+Text$\to$Text (14.8\%)}}} & 
\tikz[baseline=0.05em] \fill [color={rgb,255: red,243; green,149; blue,39}] (0,0) rectangle (0.75em,0.75em);  A-OKVQA~\cite{schwenk2022aokvqa} (17.1 K) &
\tikz[baseline=0.05em] \fill [color={rgb,255: red,244; green,161; blue,62}] (0,0) rectangle (0.75em,0.75em);   ChartQA~\cite{masry2022chartqa} (28.3 K) \\
\tikz[baseline=0.05em] \fill [color={rgb,255: red,244; green,172; blue,86}] (0,0) rectangle (0.75em,0.75em);   DocVQA~\cite{mathew2021docvqa} (39.5 K) &
\tikz[baseline=0.05em] \fill [color={rgb,255: red,244; green,184; blue,110}] (0,0) rectangle (0.75em,0.75em);  InfographicsVQA~\cite{mathew2022infographicvqa} (23.5 K) &
\tikz[baseline=0.05em] \fill [color={rgb,255: red,244; green,196; blue,133}] (0,0) rectangle (0.75em,0.75em); OK-VQA~\cite{marino2019okvqa} (9.0 K) \\
\tikz[baseline=0.05em] \fill [color={rgb,255: red,244; green,208; blue,158}] (0,0) rectangle (0.75em,0.75em); Visual7W~\cite{zhu2016visual7w} (69.8 K) \\
\hline

\makecell[l]{\parbox[c]{2.5cm}{\cellcolor[RGB]{228,84,86} \textcolor{white}{Image+Text$\to$Image (13.8\%)}}} & 
\tikz[baseline=0.05em] \fill [color={rgb,255: red,229; green,101; blue,103}] (0,0) rectangle (0.75em,0.75em); CIRR~\cite{liu2021cirr} (26.1 K) &
\tikz[baseline=0.05em] \fill [color={rgb,255: red,232; green,118; blue,120}] (0,0) rectangle (0.75em,0.75em); MSCOCO~\cite{lin2014mscoco} (100.0 K) \\
\tikz[baseline=0.05em] \fill [color={rgb,255: red,234; green,134; blue,135}] (0,0) rectangle (0.75em,0.75em); N24News~\cite{wang2022n24news} (49.0 K) \\
\hline
\makecell[l]{\parbox[c]{2.5cm}{\cellcolor[RGB]{148,116,207} \textcolor{white}{Video+Text$\to$Video (15.8\%)}}} & 
\tikz[baseline=0.05em] \fill [color={rgb,255: red,156; green,128; blue,209}] (0,0) rectangle (0.75em,0.75em);  WebVid-CoVR~\cite{ventura2024covr} (200.0 K) \\

\end{tabular}
\end{minipage}
\caption{\textbf{Instruction Tuning 1.3M.} Left: Data Distribution within Each Category. The outer circle shows the distribution of all data categories and the inner circle shows the distribution of data subsets. Right: The detailed quantities of datasets.}
\label{fig:pie_instrcut_data}
\end{figure}


%% file: tables/app-data-eval.tex
\begin{table}[htbp]
\centering
\caption{Evaluation benchmark statistics. MMEB is a comprehensive benchmark and its statistics are shown in Table \ref{tab:mmeb_statistics}. \#Text/Instruction and \#Image/Video denote the test text/instruction count and image/video pool size, respectively.}
\label{tab:eval_statistics}%
\resizebox{0.7\textwidth}{!}{
\begin{tabular}{l|cccc}
\toprule
\textbf{Benchmark} & \textbf{Query$\to$Target} & \textbf{Zero-shot} & \textbf{\#Text/Instrcution} & \textbf{\#Image/Video} \\
\midrule
\rowcolor{gray!10}\multicolumn{5}{l}{\textit{\textbf{Coarse-grained Retrieval}}}\\
\midrule
Flickr30K~\cite{plummer2015flickr30k} & T$\to$I, I$\to$T & \cmark & 1,000 & 5,000 \\
MSR-VTT~\cite{xu2016msrvtt} & T$\to$V, V$\to$T & \cmark & 1,000 & 1,000 \\
MSVD~\cite{chen2011msvd} & T$\to$V, V$\to$T & \cmark & 670 & 27,763 \\
DiDeMo~\cite{anne2017didemo} & T$\to$V, V$\to$T & \cmark & 1,004 & 1,004 \\
\midrule
\rowcolor{gray!10}\multicolumn{5}{l}{\textit{\textbf{Fine-grained Retrieval}}}\\
ShareGPT4V~\cite{plummer2015flickr30k} & T$\to$I, I$\to$T & \cmark & 1,000 & 1,000 \\
Urban1K~\cite{zhang2024longclip} & T$\to$I, I$\to$T & \cmark & 1,000 & 1,000 \\
DOCCI~\cite{onoe2024docci} & T$\to$I, I$\to$T & \cmark & 5,000 & 5,000 \\
CaRe~\cite{xu2025carebench} & T$\to$V, V$\to$T & \cmark & 1,000 & 1,000 \\
\midrule
\rowcolor{gray!10}\multicolumn{5}{l}{\textit{\textbf{Instruction-based Retrieval}}}\\
MMEB~\cite{plummer2015flickr30k} & 36 tasks & \xmark & - & - \\
WebVid-CoVR~\cite{zhang2024longclip} & T+V$\to$V & \xmark & 2,556 & 2,556 \\
\bottomrule
\end{tabular}%
}
\end{table}

%% file: tables/app-mmeb-statistics.tex
\begin{table}[!t]
\small
\centering
\caption{The statistics of MMEB: 36 datasets across 4 meta-task categories, with 20 in-distribution datasets used for training and 16 out-of-distribution datasets used exclusively for evaluation.}
\label{tab:mmeb_statistics}
\resizebox{1.0\textwidth}{!}{
\begin{tabular}{c|ccccccc}
\toprule
Meta-Task & Dataset & Query$\to$Target & Distribution Type & \#Training & \#Eval & \#Candidates \\ 
\midrule
\multirow{10}{*}{\begin{tabular}[c]{@{}c@{}}Classification \\ (10 Tasks)\end{tabular}} & ImageNet-1K       & I$\to$T         & IND & 100K & 1000 & 1000 \\  
& N24News           & I + T$\to$I     & IND &  49K & 1000 & 24   \\ 
& HatefulMemes      & I$\to$T         & IND &   8K & 1000 & 2    \\ 
& VOC2007           & I$\to$T         & IND &   8K & 1000 & 20   \\ 
& SUN397            & I$\to$T         & IND &  20K & 1000 & 397  \\ 
\cmidrule(lr){2-7}
& Place365          & I$\to$T         & OOD & -    & 1000 & 365  \\ 
& ImageNet-A        & I$\to$T         & OOD & -    & 1000 & 1000 \\ 
& ImageNet-R        & I$\to$T         & OOD & -    & 1000 & 200  \\ 
& ObjectNet         & I$\to$T         & OOD & -    & 1000 & 313  \\ 
& Country-211       & I$\to$T         & OOD & -    & 1000 & 211  \\
\midrule
\multirow{10}{*}{\begin{tabular}[c]{@{}c@{}}VQA \\ (10 Tasks)\end{tabular}} 
& OK-VQA            & I + T$\to$T     & IND &9K&1000 &1000 \\  
& A-OKVQA           & I + T$\to$T     & IND &  17K & 1000 & 1000 \\ 
& DocVQA            & I + T$\to$T     & IND &  40K & 1000 & 1000 \\ 
& InfographicVQA    & I + T$\to$T     & IND &  24K & 1000 & 1000 \\ 
& ChartQA           & I + T$\to$T     & IND &  28K & 1000 & 1000 \\  
& Visual7W          & I + T$\to$T     & IND &  70K & 1000 & 1000 \\ 
\cmidrule(lr){2-7}
& ScienceQA         & I + T$\to$T     & OOD & -   & 1000 & 1000 \\  
& VizWiz            & I + T$\to$T     & OOD & -   & 1000 & 1000 \\ 
& GQA               & I + T$\to$T     & OOD & -   & 1000 & 1000 \\ 
& TextVQA           & I + T$\to$T     & OOD & -   & 1000 & 1000 \\
\midrule
\multirow{12}{*}{\begin{tabular}[c]{@{}c@{}}Retrieval \\ (12 Tasks)\end{tabular}} 
& VisDial           & T$\to$I         & IND & 123K & 1000 & 1000 \\ 
& CIRR              & I + T$\to$I     & IND &  26K & 1000 & 1000 \\ 
& VisualNews\_t2i   & T$\to$I         & IND & 100K & 1000 & 1000 \\ 
& VisualNews\_i2t   & I$\to$T         & IND & 100K & 1000 & 1000 \\ 
& MSCOCO\_t2i       & T$\to$I         & IND & 100K & 1000 & 1000 \\ 
& MSCOCO\_i2t       & I$\to$T         & IND & 113K & 1000 & 1000 \\ 
& NIGHTS            & I$\to$I         & IND &  16K & 1000 & 1000 \\ 
& WebQA             & T$\to$I + T     & IND &  17K & 1000 & 1000 \\ 
\cmidrule(lr){2-7}
& OVEN              & I + T$\to$I + T & OOD & -   & 1000 & 1000 \\ 
& FashionIQ         & I + T$\to$I     & OOD & -   & 1000 & 1000 \\ 
& EDIS              & T$\to$I + T     & OOD & -   & 1000 & 1000 \\ 
& Wiki-SS-NQ        & T $\to$I        & OOD & -   & 1000 & 1000 \\
\midrule
\multirow{4}{*}{\begin{tabular}[c]{@{}c@{}}Visual Grounding\\ (4 Tasks)\end{tabular}} 
& MSCOCO            & I + T$\to$I     & IND & 100K & 1000 & 1000 \\  
\cmidrule(lr){2-7}
& Visual7W-Pointing & I + T$\to$I     & OOD & -   & 1000 & 1000 \\ 
& RefCOCO           & I + T$\to$I     & OOD & -   & 1000 & 1000 \\ 
& RefCOCO-Matching  & I + T$\to$I + T & OOD & -   & 1000 & 1000 \\
\bottomrule
\end{tabular}
}
\end{table}

%% file: tables/app-training-hyperparameters.tex
\begin{table}[h]
\centering
\begin{minipage}[t]{0.48\textwidth}
\caption{Training hyperparameters and computational requirements for \modelbase.}
\label{tab:training-parameters-base}
\scriptsize
\begin{tabular}{lcc}
\toprule
\multicolumn{1}{c}{\textbf{Hyperparameter}} & \model 2B & \model 7B \\
\midrule
\rowcolor{gray!10}\multicolumn{3}{c}{\textbf{\textit{Stage1: Retrieval Adaptation for} \modelbase}}\\
Training Samples & \multicolumn{2}{c}{6.4M} \\
Batch Size & 4,096 & 1,024 \\
Learning rate & \multicolumn{2}{c}{1$\times 10^{-4}$} \\
Optimizer & \multicolumn{2}{c}{AdamW} \\
Learning Rate Decay & \multicolumn{2}{c}{cosine} \\
Warmup Ratio & \multicolumn{2}{c}{0.03} \\
LoRA Rank & \multicolumn{2}{c}{8} \\
LoRA Alpha & \multicolumn{2}{c}{16} \\
Temperature $\tau$ & \multicolumn{2}{c}{0.03} \\
Epochs & \multicolumn{2}{c}{1} \\
GPU Configuration & \multicolumn{2}{c}{64$\times$A100} \\
Training Time & 7 hours & 21 hours \\
\bottomrule
\end{tabular}%
\end{minipage}
\hspace{0.01\textwidth}
\begin{minipage}[t]{0.48\textwidth}
\scriptsize
\caption{Training hyperparameters and computational requirements for \modelinstruct.}
\label{tab:training-parameters-instruct}
\begin{tabular}{lcc}
\toprule
\multicolumn{1}{c}{\textbf{Hyperparameter}} & \model 2B & \model 7B \\
\midrule
\rowcolor{gray!10}\multicolumn{3}{c}{\textbf{\textit{Stage2: Instruction Tuning for} \modelinstruct}}\\
Training Samples & \multicolumn{2}{c}{1.3M} \\
Batch Size & 4,096 & 1,024 \\
Learning rate & \multicolumn{2}{c}{2$\times 10^{-5}$} \\
Optimizer & \multicolumn{2}{c}{AdamW} \\
Learning Rate Decay & \multicolumn{2}{c}{cosine} \\
Warmup Ratio & \multicolumn{2}{c}{0.03} \\
LoRA Rank & \multicolumn{2}{c}{8} \\
LoRA Alpha & \multicolumn{2}{c}{64} \\
Temperature $\tau$ & \multicolumn{2}{c}{0.03} \\
Epochs & \multicolumn{2}{c}{1} \\
GPU Configuration & \multicolumn{2}{c}{64$\times$A100} \\
Training Time & 2 hours & 6 hours \\
\bottomrule
\end{tabular}
\end{minipage}
\end{table}

%% file: tables/app-care-full.tex
\begin{table}[htbp]
\centering
\caption{Detailed results of zero-shot performance on CaReBench~\cite{xu2025carebench}.}
\label{tab:app_fine_video}
\setlength\tabcolsep{2.2pt}
\resizebox{\textwidth}{!}{
\begin{tabular}{l|cccccc|cccccc|cccccc}
\toprule
\multirow{3}[3]{*}{\textbf{Model}} & \multicolumn{6}{c|}{\textbf{CaRe-General}} & \multicolumn{6}{c|}{\textbf{CaRe-Spatial}} & \multicolumn{6}{c}{\textbf{CaRe-Temporal}} \\
\cmidrule(lr){2-7}\cmidrule(lr){8-13}\cmidrule(lr){14-19}
& \multicolumn{3}{c}{T$\to$V} & \multicolumn{3}{c|}{V$\to$T} & \multicolumn{3}{c}{T$\to$V} & \multicolumn{3}{c|}{V$\to$T} & \multicolumn{3}{c}{T$\to$V} & \multicolumn{3}{c}{V$\to$T} \\
\cmidrule(lr){2-4}\cmidrule(lr){5-7}\cmidrule(lr){8-10}\cmidrule(lr){11-13}\cmidrule(lr){14-16}\cmidrule(lr){17-19}
& R@1 & R@5 & R@10 & R@1 & R@5 & R@10 & R@1 & R@5 & R@10 & R@1 & R@5 & R@10 & R@1 & R@5 & R@10 & R@1 & R@5 & R@10 \\
\midrule
\rowcolor{gray!10}\multicolumn{19}{l}{\textit{\textbf{CLIP-based Models}}}\\
\midrule
CLIP L/14 & 51.2 & 83.4 & 90.6 & 54.7 & 86.9 & 93.6 & 49.0 & 81.9 & 91.4 & 55.4 & 85.6 & 93.0 & 33.5 & 70.3 & 84.0 & 39.7 & 76.2 & 87.9 \\
LanguageBind & 64.3 & 91.0 & 96.3 & 59.5 & 88.0 & 95.0 & 64.7 & 90.8 & 96.8 & 61.0 & 87.2 & 94.5 & 39.8 & 77.3 & 90.5 & 42.2 & 77.6 & 91.7 \\
Long-CLIP L/14 & 62.7 & 88.8 & 95.7 & 60.3 & 88.8 & 94.9 & 65.6 & 90.9 & 96.0 & 61.0 & 88.3 & 94.4 & 33.2 & 68.8 & 81.6 & 34.5 & 71.9 & 86.6 \\
\text{InternVideo2}$_\text{stage2}$ 1B & 72.5 & 93.7 & 97.3 & 69.5 & 94.6 & 97.8 & 72.4 & 94.2 & 97.4 & 62.7 & 90.5 & 95.9 & 46.0 & 80.8 & 91.9 & 46.6 & 82.5 & 92.5 \\
\midrule
\rowcolor{gray!10}\multicolumn{19}{l}{\textit{\textbf{LMM-based Models}}}\\
\midrule
LLaVA-NV 7B$^\dagger$ & 66.9 & 89.4 & 96.0 & 62.7 & 89.2 & 95.4 & 68.0 & 92.0 & 96.2 & 65.0 & 90.0 & 95.9 & 43.3 & 76.9 & 88.9 & 40.1 & 75.4 & 88.7 \\
MiniCPM-V 2.6$^\dagger$ & 71.0 & 92.2 & 97.0 & 69.3 & 92.8 & 97.1 & 71.7 & 93.6 & 98.0 & 67.6 & 92.3 & 97.7 & 50.5 & 82.9 & 92.1 & 46.1 & 80.9 & 93.3 \\
InternVL2 8B$^\dagger$ & 72.1 & 92.6 & 96.8 & 73.6 & 93.4 & 97.4 & 76.1 & 94.1 & 97.6 & 74.3 & 94.5 & 97.6 & 48.1 & 76.8 & 89.0 & 47.6 & 78.2 & 90.3 \\
Tarsier 7B$^\dagger$ & 71.0 & 93.8 & 97.8 & 70.6 & 94.2 & 98.0 & 70.2 & 94.0 & 98.2 & 67.4 & 93.5 & 97.4 & 50.1 & 84.1 & 92.8 & 50.0 & 84.7 & 94.9 \\
Qwen2-VL 7B$^\dagger$ & 76.6 & 95.3 & \second{98.7} & 77.4 & 95.6 & 98.7 & 78.2 & 95.5 & 98.5 & 75.4 & 95.0 & 98.1 & \second{51.9} & \second{84.8} & \best{94.9} & 52.7 & 85.4 & \second{95.2} \\
CaRe 7B & 77.0 & \second{95.6} & \second{98.7} & 79.0 & \second{96.8} & \second{99.1} & 76.8 & \second{96.3} & \second{98.7} & \second{78.1} & \second{95.8} & \second{99.3} & 50.7 & \best{85.3} & \second{94.4} & \second{53.4} & \second{86.3} & 94.0 \\
\midrule
\modelbase 2B & \second{78.1} & 95.5 & 97.8 & \second{80.8} & 96.4 & 98.7 & \second{79.6} & 95.4 & 98.3 & 78.0 & 95.4 & 97.9 & 45.3 & 77.6 & 89.9 & 50.0 & 83.6 & 92.6 \\
\modelbase 7B & \best{86.0} & \best{96.8} & \best{98.9} & \best{86.9} & \best{98.3} & \best{99.7} & \best{86.5} & \best{96.9} & \best{99.2} & \best{84.8} & \best{98.0} & \best{99.4} & \best{52.4} & 82.5 & 92.2 & \best{55.4} & \best{86.5} & \best{95.3} \\
\bottomrule
\end{tabular}%
}
\end{table}

%% file: tables/app-mmeb-full.tex
\begin{table}[t]
\centering
\caption{Detailed results on the MMEB benchmark~\cite{jiang2024vlm2vec}. We average the scores in each meta-task.}
\label{tab:app_mmeb_full}%
\resizebox{\textwidth}{!}{
\begin{tabular}{lccccccc}
\toprule
\multirow{2}[2]{*}{\textbf{Model}} & \multicolumn{4}{c}{\textbf{Per Meta-Task Score}} & \multicolumn{3}{c}{\textbf{Average Score}} \\ 
\cmidrule(lr){2-5}\cmidrule(lr){6-8}
& Classification & VQA & Retrieval & Grounding & IND & OOD & Overall \\ \midrule
\#Datasets $\rightarrow$ & 10 & 10 & 12 & 4 & 20 & 16 & 36 \\
\midrule
\rowcolor{gray!10}\multicolumn{8}{c}{\textit{\textbf{Zero-shot Setting}}}\\
\midrule
CLIP~\cite{radford2021clip}                & 42.8 & 9.1  & 53.0 & 51.8 & 37.1 & 38.7 & 37.8 \\
BLIP2~\cite{li2023blip2}                   & 27.0 & 4.2  & 33.9 & 47.0 & 25.3 & 25.1 & 25.2 \\
SigLIP~\cite{zhai2023siglip}               & 40.3 & 8.4  & 31.6 & 59.5 & 32.3 & 38.0 & 34.8 \\
OpenCLIP~\cite{cherti2023openclip}         & 47.8 & 10.9 & 52.3 & 53.3 & 39.3 & 40.2 & 39.7 \\
Magiclens~\cite{zhang2024magiclens}        & 38.8 & 8.3  & 35.4 & 26.0 & 31.0 & 23.7 & 27.8 \\
E5-V 8B (LLaVA-NeXT)~\cite{jiang2024e5v}   & 21.8 & 4.9  & 11.5 & 19.0 & 14.9 & 11.5 & 13.3 \\
MMRet-MLLM 7B (LLaVA-1.6)~\cite{zhou2024megapairs} & 47.2 & 18.4 & 56.5 & 62.2 & 43.5 & 44.3 & 44.0 \\
mmE5 11B (Llama-3.2-Vision)~\cite{chen2025mme5}   & 60.6 & 55.7 & 54.7 & 72.4 & 57.2 & 62.9 & 58.6 \\
\midrule
\rowcolor{gray!10}\multicolumn{8}{c}{\textit{\textbf{Partially Supervised Finetuning Setting (Finetuning on M-BEIR)}}}\\
\midrule
UniIR (BLIP\_FF)~\cite{wei2024uniir}            & 42.1 & 15.0 & 60.1 & 62.2 & 44.7 & 40.4 & 42.8 \\
UniIR (CLIP\_SF)~\cite{wei2024uniir}            & 70.5 & 16.2 & 61.8 & 65.3 & 47.1 & 41.7 & 44.7 \\
MM-Embed 7B (LLaVA-1.6)~\cite{lin2024mmembed}   & 48.1 & 32.3 & 63.8 & 57.8 & -    & -    & 50.0 \\
GME 2B (Qwen2-VL)~\cite{zhang2024gme}           & 56.9 & 41.2 & 67.8 & 53.4 & -    & -    & 55.8 \\
\midrule
\rowcolor{gray!10}\multicolumn{8}{c}{\textit{\textbf{Supervised Finetuning Setting (Finetuning on MMEB)}}}\\
\midrule
CLIP~\cite{radford2021clip}                & 55.2 & 19.7 & 53.2 & 62.2 & 47.6 & 42.8 & 45.4 \\
OpenCLIP~\cite{cherti2023openclip}         & 56.0 & 21.9 & 55.4 & 64.1 & 50.5 & 43.1 & 47.2 \\
E5-V 4.2B (Phi3.5-V)~\cite{jiang2024e5v}   & 39.1 & 9.6  & 38.0 & 57.6 & 33.1 & 31.9 & 32.6 \\
E5-V 7B (LLaVA-1.6)~\cite{jiang2024e5v}    & 39.7 & 10.8 & 39.4 & 60.2 & 34.2 & 33.4 & 33.9 \\
VLM2Vec 4.2B (Phi-3.5-V)~\cite{jiang2024vlm2vec}         & 54.8 & 54.9 & 62.3 & 79.5 & 66.5 & 52.0 & 60.1 \\
VLM2Vec 7B (LLaVA-1.6)~\cite{jiang2024vlm2vec} & 61.2 & 49.9 & 67.4 & 86.1 & 67.5 & 57.1 & 62.9 \\
VLM2Vec 2B (Qwen2-VL)~\cite{jiang2024vlm2vec}  & 59.0 & 49.4 & 65.4 & 73.4 & 66.0 & 52.6 & 60.1 \\
VLM2Vec 7B (Qwen2-VL)~\cite{jiang2024vlm2vec}  & 62.6 & 57.8 & 69.9 & 81.7 & 72.2 & 57.8 & 65.8 \\
MMRet-MLLM 7B (LLaVA-1.6)~\cite{zhou2024megapairs}       & 56.0 & 57.4 & 69.9 & 83.6 & 68.0 & 59.1 & 64.1 \\
UniME  4.2B (Phi3.5-V)~\cite{gu2025unime}           & 54.8 & 55.9 & 64.5 & 81.8 & 68.2 & 52.7 & 64.2 \\
UniME 7B (LLaVA-1.6)~\cite{gu2025unime}             & 60.6 & 52.9 & 67.9 & 85.1 & 68.4 & 57.9 & 66.6 \\
CAFe 0.5B (LLaVA-OneVision)~\cite{yu2025cafe}       & 59.1 & 49.1 & 61.0 & 83.0 & 64.3 & 53.7 & 59.6 \\
CAFe 7B (LLaVA-OneVision)~\cite{yu2025cafe}         & 65.2 & \best{65.6} & 70.0 & \best{91.2} & \best{75.8} & 62.4 &  \second{69.8} \\
mmE5 11B (Llama-3.2-Vision)~\cite{chen2025mme5} & \second{67.6} & 62.8 & 70.9 & \second{89.7} & 72.3 & \best{66.7} & \second{69.8} \\
IDMR 8B (InternVL2.5)~\cite{liu2025idmr}   & 58.3 & 58.6 & 68.7 & 85.6 & 70.5 & 57.9 & 64.9 \\
IDMR 26B (InternVL2.5)~\cite{liu2025idmr}  & 66.3 & 61.9 & \second{71.1} & 88.6 & 73.4 & 63.9 & 69.2 \\
\midrule
\modelinstruct 2B (Qwen2-VL) & 63.2 & 55.9 & 65.4 & 75.6 & 65.8 & 60.1 & 63.3 \\
\modelinstruct 7B (Qwen2-VL) & \best{68.3} & \second{65.1} & \best{71.6} & 84.8 & \second{73.6} & \second{66.3} & \best{70.3} \\
\bottomrule
\end{tabular}%
}
\end{table}


%% file: tables/app-coarse-video-text-full.tex
\begin{table}[htbp]
\centering
\caption{Detailed results of zero-shot performance on MSR-VTT~\cite{xu2016msrvtt}, MSVD~\cite{chen2011msvd}, and DiDeMo~\cite{anne2017didemo}.}
\label{tab:app_coarse_video}%
\setlength\tabcolsep{2pt}
\resizebox{\textwidth}{!}{
\begin{tabular}{l|cccccc|cccccc|cccccc}
\toprule
\multirow{3}[3]{*}{\textbf{Model}} & \multicolumn{6}{c|}{\textbf{MSR-VTT}} & \multicolumn{6}{c|}{\textbf{MSVD}} & \multicolumn{6}{c}{\textbf{DiDeMo}} \\
\cmidrule(lr){2-7}\cmidrule(lr){8-13}\cmidrule(lr){14-19}
& \multicolumn{3}{c}{T$\to$V} & \multicolumn{3}{c|}{V$\to$T} & \multicolumn{3}{c}{T$\to$V} & \multicolumn{3}{c|}{V$\to$T} & \multicolumn{3}{c}{T$\to$V} & \multicolumn{3}{c}{V$\to$T} \\
\cmidrule(lr){2-4}\cmidrule(lr){5-7}\cmidrule(lr){8-10}\cmidrule(lr){11-13}\cmidrule(lr){14-16}\cmidrule(lr){17-19}
& R@1 & R@5 & R@10 & R@1 & R@5 & R@10 & R@1 & R@5 & R@10 & R@1 & R@5 & R@10 & R@1 & R@5 & R@10 & R@1 & R@5 & R@10 \\
\midrule
\rowcolor{gray!10}\multicolumn{19}{l}{\textit{\textbf{CLIP-based Models}}}\\
\midrule
CLIP L/14~\cite{radford2021clip} & 36.7 & 58.8 & 68.0 & 32.8 & 54.7 & 66.2 & 41.1 & 68.8 & 77.5 & 68.1 & 85.5 & 91.8 & 24.1 & 48.0 & 58.2 & 23.8 & 44.9 & 54.0 \\
Long-CLIP L/14~\cite{zhang2024longclip} & 40.9 & 65.5 & 74.6 & 36.2 & 62.2 & 71.5 & 46.5 & 73.5 & 82.0 & 69.3 & 86.0 & 90.3 & 32.4 & 56.2 & 65.2 & 28.5 & 54.1 & 64.7 \\
InternVideo~\cite{wang2022internvideo}  & 40.7 & -    & -    & 39.6 & -    & -    & 43.4 & -    & -    & 67.6 & -    & -    & 31.5 & -    & -    & 33.5 & -    & -    \\
LanguageBind~\cite{zhu2023languagebind} & 42.1 & 65.9 & 75.5 & 40.1 & 65.4 & 73.9 & 50.0 & 77.7 & 85.6 & 75.1 & 90.0 & 94.2 & 35.6 & 63.6 & 71.7 & 35.6 & 62.8 & 71.8 \\
ViCLIP~\cite{wang2023internvid} & 42.4 & -    & -    & 41.3 & -    & -    & 49.1 & -    & -    & 75.1 & -    & -    & 18.4 & -    & -    & 27.9 & -    & -    \\
\midrule
\rowcolor{gray!10}\multicolumn{19}{l}{\textit{\textbf{LMM-based Models}}}\\
\midrule
VLM2Vec~\cite{jiang2024vlm2vec} & 46.8 & 71.1 & 80.0 & -    & -    & -    & 52.9 & 80.1 & 87.0 & -    & -    & -    & -    & -    & -    & -    & -    & -    \\
LamRA 7B~\cite{liu2024lamra} & 44.7 & 68.6 & 78.6 & -    & -    & -    & 52.4 & 79.8 & 87.0 & -    & -    & -    & -    & -    & -    & -    & -    & -    \\
CaRe 7B~\cite{xu2025carebench} & 43.9 & 67.0 & 75.7 & 41.7 & 68.1 & 76.2 & 52.6 & 79.2 & 86.6 & 74.6 & 87.9 & 92.4 & 41.4 & 68.5 & 77.1 & 39.1 & 66.0 & 75.8 \\
\midrule
\modelbase 2B       & 43.8 & 67.6 & 76.8 & 41.7 & 66.2 & 75.9 & 50.0 & 77.8 & 85.1 & 73.1 & 90.7 & 94.3 & 37.9 & 64.1 & 73.7 & 26.4 & 37.5 & 63.5 \\
\modelbase 7B       & 46.5 & 69.4 & 78.0 & 45.2 & 70.3 & 79.3 & 50.4 & 78.2 & 86.4 & 76.1 & 91.3 & 94.6 & 43.5 & 69.6 & 77.3 & 40.3 & 68.7 & 78.1 \\
\bottomrule
\end{tabular}%
}
\end{table}

%% file: sections/limitations.tex
\section*{Limitations}
\label{sec:limitations}

While \model demonstrates superior performance across text, image, and video modalities, audio emerges as another potential modality with the evolution of social media. Our exploration reveals that balancing multiple modalities remains challenging, suggesting the need for further investigation into modality expansion. Additionally, while comprehensive benchmarks exist for image-text retrieval, developing a unified benchmark that encompasses text, image, video, and potentially audio modalities represents a valuable future direction.

%% file: tables/app-mmeb-per-task.tex
\begin{table}[]
\centering
\caption{The detailed results of the baselines and our \model on MMEB, which includes 20 in-distribution (IND) datasets and 16 out-of-distribution (OOD) datasets. The out-of-distribution datasets are highlighted with a yellow background in the table. For each model, we report its best variant with fully available performance metrics, including VLM2Vec 7B (LLaVA-1.6)~\cite{jiang2024vlm2vec}, MMRet 7B (LLaVA-1.6)~\cite{zhou2024megapairs}, UniME 7B (LLaVA-1.6)~\cite{gu2025unime}, mmE5 11B (Llama-3.2-Vision)~\cite{chen2025mme5}, and IDMR 26B (InternVL2.5)~\cite{liu2025idmr}. Superior versions might exist but are excluded due to incomplete score reporting.}
\resizebox{1.0\textwidth}{!}{
\begin{tabular}{lcccccccc}
\toprule
\rowcolor{gray!30}  & \textbf{CLIP} & \textbf{VLM2Vec} & \textbf{MMRet} & \textbf{UniME} & \textbf{mmE5} & \textbf{IDMR} & \textbf{\model 2B} & \textbf{\model 7B} \\
\midrule
\rowcolor{orange!30} \textbf{Classification (10 tasks)} & & & & & & & &\\
ImageNet-1K          & 55.8 & 74.5 & 58.8 & 71.3 & 77.8 & 80.6  & 77.6 & 80.2 \\
N24News              & 34.7 & 80.3 & 71.3 & 79.5 & 81.7 & 81.6  & 66.8 & 80.3 \\
HatefulMemes         & 51.1 & 67.9 & 53.7 & 64.6 & 64.2 & 72.3  & 57.4 & 67.1 \\
VOC2007              & 50.7 & 91.5 & 85.0 & 90.4 & 91.0 & 92.7  & 85.0 & 84.9 \\
SUN397               & 43.4 & 75.8 & 70.0 & 75.9 & 77.7 & 78.8  & 75.2 & 78.7 \\
\rowcolor{yellow!15} Place365  & 28.5 & 44.0 & 43.0 & 45.6 & 43   & 38.9  & 41.3 & 44.5 \\
\rowcolor{yellow!15} ImageNet-A & 25.5 & 43.6 & 36.1 & 45.5 & 56.3 & 63.6  & 48.3 & 59.2 \\
\rowcolor{yellow!15} ImageNet-R & 75.6 & 79.8 & 71.6 & 78.4 & 86.3 & 84    & 88.8 & 90.5 \\
\rowcolor{yellow!15} ObjectNet  & 43.4 & 39.6 & 55.8 & 36.4 & 62.5 & 50.5  & 66.6 & 68.1 \\
\rowcolor{yellow!15} Country-211 & 19.2 & 14.7 & 14.7 & 18.7 & 35.4 & 20.3  & 24.8 & 29.5 \\
\textit{All Classification} & 42.8 & 61.2 & 56.0 & 60.6 & 67.6 & 66.3 & 63.2 & 68.3 \\
\midrule

\rowcolor{blue!30} \textbf{VQA (10 tasks)} & & & & & & & & \\
OK-VQA               & 7.5  & 69.0 & 73.3 & 68.3 & 67.6 & 71.0  & 57.3 & 67.1 \\
A-OKVQA              & 3.8  & 54.4 & 56.7 & 58.7 & 56.1 & 59.2  & 46.0 & 58.0 \\
DocVQA               & 4.0  & 52.0 & 78.5 & 67.6 & 90.3 & 75.1  & 87.0 & 92.7 \\
InfographicsVQA      & 4.6  & 30.7 & 39.3 & 37.0 & 56.5 & 44.6  & 52.8 & 71.3 \\
ChartQA              & 1.4  & 34.8 & 41.7 & 33.4 & 50.5 & 64.6  & 45.7 & 63.2 \\
Visual7W             & 4.0  & 49.8 & 49.5 & 51.7 & 51.9 & 54.9  & 47.7 & 54.9 \\
\rowcolor{yellow!15} ScienceQA  & 9.4  & 42.1 & 45.2 & 40.5 & 55.8 & 54.7 & 41.1 & 51.2 \\
\rowcolor{yellow!15} VizWiz     & 8.2  & 43.0 & 51.7 & 42.7 & 52.8 & 47.1 & 49.8 & 53.4 \\
\rowcolor{yellow!15} GQA        & 41.3 & 61.2 & 59.0 & 63.6 & 61.7 & 71.0 & 53.0 & 56.8 \\
\rowcolor{yellow!15} TextVQA    & 7.0  & 62.0 & 79.0 & 65.2 & 83.3 & 77.0 & 78.9 & 82.3 \\
\textit{All VQA}      & 9.1  & 49.9 & 57.4 & 52.9 & 62.6 & 61.9 & 55.9 & 65.1 \\
\midrule

\rowcolor{green!30} \textbf{Retrieval (12 tasks)} & & & & & & & & \\
VisDial              & 30.7 & 80.9 & 83.0 & 79.7 & 74.1 & 81.5 & 70.0 & 80.5 \\
CIRR                 & 12.6 & 49.9 & 61.4 & 52.2 & 54.7 & 57.6 & 43.5 & 51.6 \\
VisualNews\_t2i      & 78.9 & 75.4 & 74.2 & 74.8 & 77.6 & 78.5 & 70.6 & 79.3 \\
VisualNews\_i2t      & 79.6 & 80.0 & 78.1 & 78.8 & 83.3 & 80.6 & 74.1 & 82.4 \\
MSCOCO\_t2i          & 59.5 & 75.7 & 78.6 & 74.9 & 76.4 & 79.1 & 73.6 & 78.2 \\
MSCOCO\_i2t          & 57.7 & 73.1 & 72.4 & 73.8 & 73.2 & 75.4 & 71.1 & 74.3 \\
NIGHTS               & 60.4 & 65.5 & 68.3 & 66.2 & 68.3 & 68.6 & 65.1 & 66.0 \\
WebQA                & 67.5 & 87.6 & 90.2 & 89.8 & 88.0 & 89.0 & 85.0 & 87.0 \\
\rowcolor{yellow!15} FashionIQ  & 11.4 & 16.2 & 54.9 & 16.5 & 28.8 & 21.0 & 18.1 & 26.3 \\
\rowcolor{yellow!15} Wiki-SS-NQ & 55.0 & 60.2 & 24.9 & 66.6 & 65.8 & 66.9 & 65.0 & 72.2 \\
\rowcolor{yellow!15} OVEN       & 41.1 & 56.5 & 87.5 & 55.7 & 77.5 & 67.4 & 65.9 & 73.1 \\
\rowcolor{yellow!15} EDIS       & 81.0 & 87.8 & 65.6 & 86.2 & 83.7 & 87.6 & 82.2 & 88.3 \\
\textit{All Retrieval} & 53.0 & 67.4 & 69.9 & 67.9 & 71.0 & 71.1 & 65.4 & 71.6 \\
\midrule

\rowcolor{purple!30} \textbf{Visual Grounding (4 tasks)} & & & & & & & & \\
MSCOCO         & 33.8 & 80.6 & 76.8 & 76.5 & 53.7 & 81.5 & 64.6 & 73.9 \\
\rowcolor{yellow!15} RefCOCO  & 56.9 & 88.7 & 89.8 & 89.3 & 92.7 & 91.7 & 79.5 & 89.2 \\
\rowcolor{yellow!15} RefCOCO-matching  & 61.3 & 84.0 & 90.6 & 90.6 & 88.8 & 88.1 & 84.5 & 90.1 \\
\rowcolor{yellow!15} Visual7W-pointing & 55.1 & 90.9 & 77.0 & 84.1 & 92.3 & 93.1 & 73.9 & 86.1 \\
\textit{All Visual Grounding} & 51.8 & 86.1 & 83.6 & 85.1 & 89.6 & 88.6 & 75.6 & 84.8 \\
\midrule

\rowcolor{cyan!15} \textbf{Final Score (36 tasks)} & & & & &  & & & \\
All           & 37.8 & 62.9 & 64.1 & 66.6 & 69.8 & 69.2 & 63.3 & 70.3 \\
All IND       & 37.1 & 67.5 & 59.1 & 68.4 & 72.3 & 73.4 & 65.8 & 73.6 \\
All OOD       & 38.7 & 57.1 & 68.0 & 57.9 & 66.7 & 63.4 & 60.1 & 66.3 \\

\bottomrule
\end{tabular}
}
\label{tab:app_mmeb_per_task}
\end{table}